\documentclass[10pt,journal,compsoc]{IEEEtran}
\usepackage{ifpdf}

\ifCLASSOPTIONcompsoc
  \usepackage[nocompress]{cite}
\else
  \usepackage{cite}
\fi
\ifCLASSINFOpdf
  \usepackage[pdftex]{graphicx}
\else
   \usepackage[dvips]{graphicx}
  \graphicspath{{../eps/}}
\fi
\usepackage{amsmath}
\usepackage[justification=raggedright ]{caption}

\usepackage{algorithmic}

\usepackage{array}

\usepackage{stfloats}
\usepackage{multirow}
\usepackage{color}
\usepackage{url}

\usepackage{amsfonts}

\begin{document}
%

\title{StrokeGAN+: Few-Shot Semi-Supervised Chinese Font Generation with Stroke Encoding}
%
%
%
%

\author{Jinshan~Zeng,
        ~Yefei~Wang,
        ~Qi~Chen,
        ~Yunxin~Liu,
        ~Mingwen~Wang,
        ~and Yuan Yao
\IEEEcompsocitemizethanks{\IEEEcompsocthanksitem
J. Zeng, Y. Wang, Q. Chen, Y. Liu and M. Wang are with the School of Computer and Information Engineering, Jiangxi Normal University, Nanchang, China, 330022. E-mail: jinshanzeng@jxnu.edu.cn, fei@jxnu.edu.cn, chenqi970226@gmail.com, 201841600023@jxnu.edu.cn, mwwang@jxnu.edu.cn.
\IEEEcompsocthanksitem Y. Yao is with Department of Mathematics, Hong Kong University of Science and Technology, Hong Kong. Email: yuany@ust.hk. (Corresponding author)
}

}

\IEEEtitleabstractindextext{%
\begin{abstract}
The generation of Chinese fonts has a wide range of applications. The currently predominated methods are mainly based on deep generative models, especially the generative adversarial networks (GANs). However, existing GAN-based models usually suffer from the well-known mode collapse problem. When mode collapse happens, the kind of GAN-based models will be failure to yield the correct fonts. To address this issue, we introduce a one-bit stroke encoding and a few-shot semi-supervised scheme (i.e., using a few paired data as semi-supervised information) to explore the local and global structure information of Chinese characters respectively, motivated by the intuition that strokes and characters directly embody certain local and global modes of Chinese characters. Based on these ideas, this paper proposes an effective model called \textit{StrokeGAN+}, which incorporates the stroke encoding and the few-shot semi-supervised scheme into the CycleGAN model. The effectiveness of the proposed model is demonstrated by amounts of experiments. Experimental results show that the mode collapse issue can be effectively alleviated by the introduced one-bit stroke encoding and few-shot semi-supervised training scheme, and that the proposed model outperforms the state-of-the-art models in fourteen font generation tasks in terms of four important evaluation metrics and the quality of generated characters. Besides CycleGAN, we also show that the proposed idea can be adapted to other existing models to improve their performance. The effectiveness of the proposed model for the zero-shot traditional Chinese font generation is also evaluated in this paper.
\end{abstract}

}

\maketitle

\IEEEdisplaynontitleabstractindextext

%
\IEEEpeerreviewmaketitle

\IEEEraisesectionheading{\section{Introduction}\label{sec:introduction}}

%
%
%
%

\begin{figure}[htbp]
	\centering
	\begin{minipage}{1\linewidth}
		\centering
		\includegraphics[width=1\linewidth]{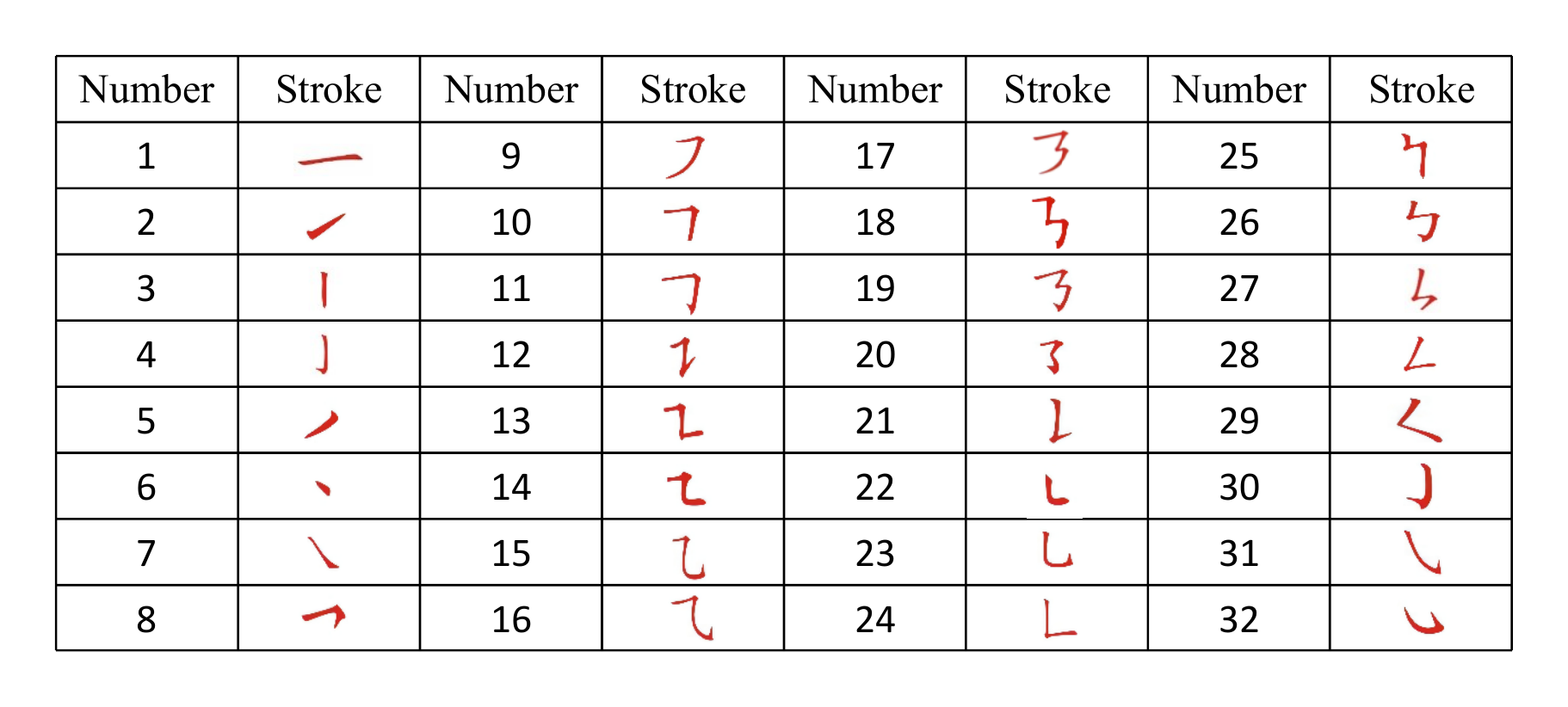}
		\centerline{{\small (a) 32 basic strokes}}
	\end{minipage}
	\begin{minipage}{0.312\linewidth}
		\centering
		\includegraphics[width=.88\linewidth]{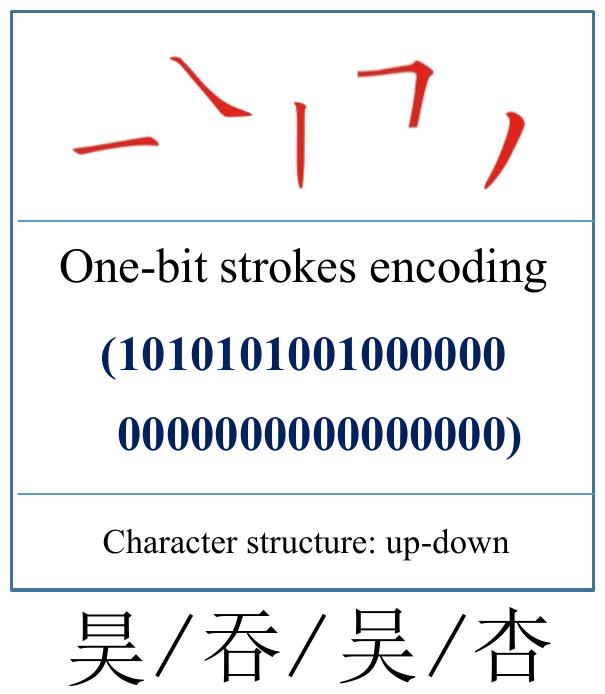}
		\begin{center}
\small (b) Stroke encoding
\end{center}
	\end{minipage}
		\begin{minipage}{0.325\linewidth}
		\centering
		\includegraphics[width=0.5\linewidth]{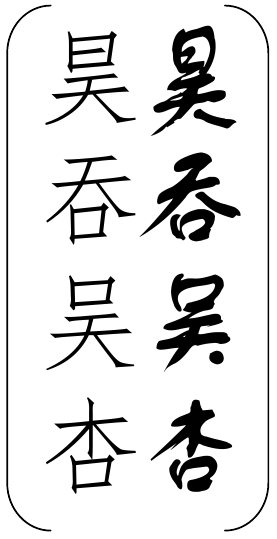}
				\begin{center}
\small (c) Paired data
\end{center}
	\end{minipage}
	\begin{minipage}{0.325\linewidth}
		\centering
		\includegraphics[width=0.5\linewidth]{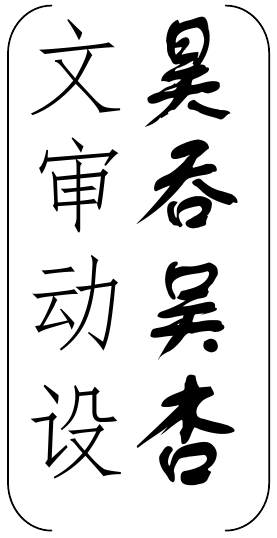}
				\begin{center}
\small (d) Unpaired data
\end{center}
	\end{minipage}
		\caption{(a) 32 basic strokes that make up Chinese characters.  (b)  Examples of of Chinese characters with the same one-bit stroke encodings and character structures. (c) Examples of paried data (left: Fangsong, right: Shuti). (d) Examples of unpaired data.}
		\label{Fig:stroke}
\end{figure}

\IEEEPARstart{I}{n} recent years, the generation of Chinese fonts has received a large amount of attention due to its central role in a wide range of applications such as the art font design \cite{wang2022aesthetic,liu2022}, personalized font design \cite{lin2014font}, and calligraphy font generation \cite{qian2007towards,chen2011chinese}. With the emergence of the generative adversarial network (GAN) \cite{goodfellow2014generative}, the kind of models based on GAN become the mainstream for the generation of Chinese fonts. These models can be generally divided into two categories, i.e., the class of models based on the paired data (that is, there is a one-to-one correspondence between the source and target font domains as shown in Figure \ref{Fig:stroke}(c)) \cite{lyu2017auto,tian2017zi2zi,wu2020calligan,yuan2022se,Song2022lffont}, and the class of models based on the unpaired data (that is, such a one-to-one correspondence is not required as shown in Figure \ref{Fig:stroke}(d)) \cite{chang2018generating,li2019improving,jiang2019scfont,lin2020chinese,xie2021dg}.

In the pioneer work \cite{lyu2017auto}, an auto-encoder guided GAN model was proposed for Chinese calligraphy synthesis through treating this problem as an image-to-image translation problem based on paired samples. Starting from this perspective, in the well-known project \cite{tian2017zi2zi}, the developers adopted the Pix2Pix model \cite{isola2017image} emerged in the research field of image style transfer to the task of Chinese font generation. In \cite{wu2020calligan}, the authors proposed an effective model called \textit{CalliGAN} by incorporating some kinds of component information such as strokes of Chinese characters, where several auxiliary network modules were introduced to extract these kinds of component information. In the recent paper \cite{yuan2022se}, the authors proposed a novel GAN-based image translation model by integrating the skeleton information for the generation of Chinese brush handwriting font. In \cite{Song2022lffont}, the authors proposed a font generation method that learns localized styles, namely component-wise style representations. Although the performance of this kind of model based on paired data in the literature \cite{lyu2017auto,tian2017zi2zi,wu2020calligan,yuan2022se,Song2022lffont} is impressive, the collection of extensive paired samples is generally labour-intensive and expensive. Particularly, in some font generation tasks such as the generation of ancient calligraphy fonts \cite{qian2007towards,chen2011chinese}, it is hard to yield extensive paired samples.

\begin{figure*}[!t]
\begin{center}
	\center{\includegraphics[width=18cm]  {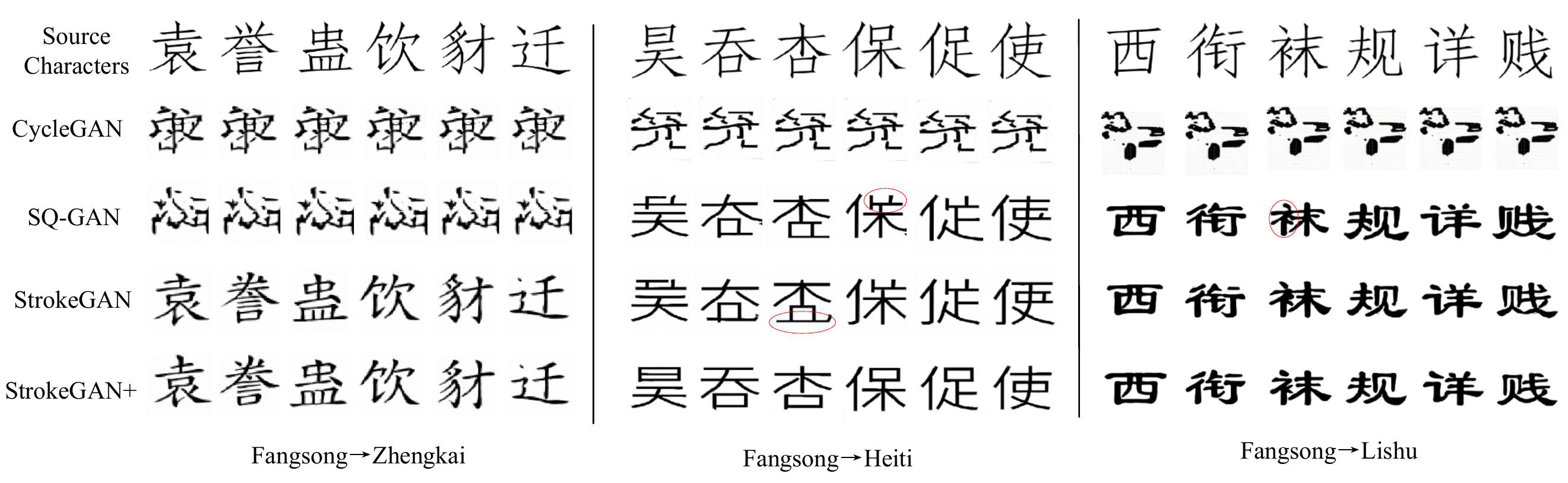}}
	\caption{The performance of the proposed \textit{StrokeGAN+} in reducing mode collapse over three generation tasks in comparison to the baselines including CycleGAN \cite{chang2018generating}, SQ-GAN \cite{zeng2022Square-BlockGAN} and StrokeGAN \cite{zeng2021strokegan}. From left to right of the tasks: Fangsong font to Zhengkai font, Fangsong to Heiti, and Fangsong to Lishu. The second row to the fifth row represent the performance of different models: the original CycleGAN encounters mode collapse over these three generation tasks; the CycleGAN with square-block translations (SQ-GAN) \cite{zeng2022Square-BlockGAN} still suffers from the mode collapse over the task \{Fangsong$\rightarrow$Zhengkai\}; the CycleGAN only with one-bit stroke encodings (StrokeGAN) sometimes generates characters with some flaws; the proposed StrokeGAN+ can effectively address the mode collapse issue and reduce these flaws by simultaneously integrating the one-bit stroke encodings and few-shot semi-supervised training scheme.
	}
	\label{Fig:mode collapse and stroke missing fig}
	\end{center}
\end{figure*}

To address this issue, many models based on unpaired data have been suggested in the literature \cite{chang2018generating,li2019improving,jiang2019scfont,lin2020chinese,xie2021dg}. In \cite{chang2018generating}, the authors exploited the well-known CycleGAN model\cite{zhu2017unpaired} developed in the field of style transfer for Chinese font generation based on the unpaired  data. In \cite{li2019improving}, the authors proposed an effective model by using graph matching for the calligraphy character generation. In \cite{jiang2019scfont}, the authors proposed a structure-guided deep generative model for the generation of Chinese fonts (there dubbed \textit{SCFont}) through integrating the domain knowledge of Chinese characters such as writing trajectory and skeleton. In \cite{xie2021dg}, a novel deformable generative model called \textit{DG-Font} was proposed for the generation of Chinese font based on the unpaired data.  Despite the effectiveness of these models, the kind of unsupervised models (say, CycleGAN \cite{chang2018generating}) usually suffers from the well-known mode collapse issue \cite{goodfellow2014generative}, which results in poor performance. Some examples are shown in the second row of Figure \ref{Fig:mode collapse and stroke missing fig}.

To tackle the mode collapse problem, the recent paper \cite{zeng2022Square-BlockGAN} incorporated a square-block geometric transformation based self-supervised scheme into the CycleGAN model to capture some simple spatial structures of Chinese characters (e.g., up-bottom, left-right structures). Their model is dubbed square-block transformation based GAN (called SQ-GAN for short henceforth). However, such kind of global features, i.e., several typical spatial structures of Chinese characters are generally insufficient to thoroughly solve this issue as shown in the third row of Figure \ref{Fig:mode collapse and stroke missing fig}, where SQ-GAN may still suffer from the mode collapse issue in some generation tasks, mainly due to many Chinese characters with the same spatial structures as depicted in Figure \ref{Fig:stroke}(b).

Inspired by the observation that the local and global structure information of Chinese characters is closely related to their strokes and counterparts in the target font domain respectively as shown in Figure \ref{Fig:stroke} (a)-(c), and the fact that a few paired samples are generally easy to access in practice, this paper aims to suggest more effective strategies for alleviating the mode collapse in the generation of Chinese fonts by investigating the following research questions:

\begin{enumerate}
    \item[(Q1)] Can the use of stroke information alleviate the mode collapse?
    \item[(Q2)] Can the use of few-shot paired data further improve the performance?
\end{enumerate}

\begin{figure}[!t]
\begin{center}
	\center{\includegraphics[width=8cm]  {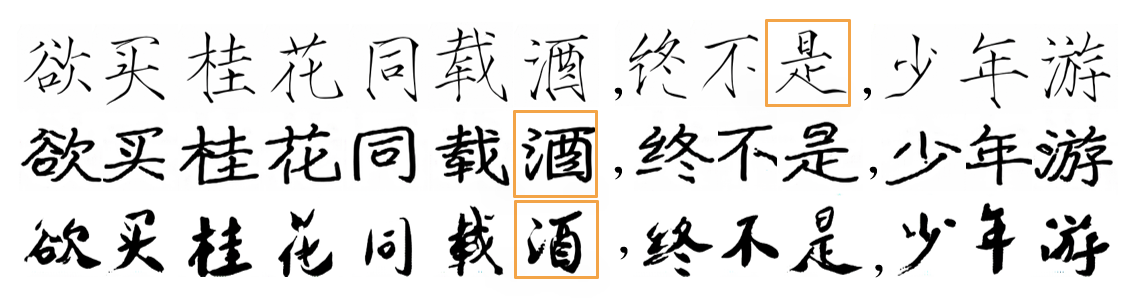}}
	\caption{An identical text string is rendered by the proposed model in three different font styles, i.e., Shoujin (1st row),  Xingkai (2nd row) and Caoshu (3rd row). The characters marked by the yellow boxes are real characters with the target font styles, while the other characters are generated by the proposed model.}
	\label{Fig:results-presentation}
	\end{center}
\end{figure}

Since each Chinese character is composed of 32 basic strokes as shown in Figure \ref{Fig:stroke}(a), we introduce a simply one-bit stroke encoding of 32 dimensions to embody the local information of a Chinese character (some examples can be found in Figure \ref{Fig:stroke}(b)) and incorporate it into the CycleGAN model as certain supervision information to guide the model to capture the local structure information of Chinese characters. Noticing that some characters have the same stroke encodings but different spatial structures, we use a few paired data as few-shot semi-supervised information to explore the global spatial structure information to further distinguish them. To fully exploit these information, we introduce a stroke reconstruction loss and a few-shot semi-supervised (FS$^3$) loss to guide the networks to particularly capture these local and global structures of Chinese characters. The major contributions of this paper can be summarized as follows:
\begin{enumerate}
    \item This paper proposes an effective model called \textit{StrokeGAN+} for the generation of Chinese fonts by incorporating both local and global structure information into CycleGAN \cite{chang2018generating} to alleviate the well-known mode collapse issue as well as improve the performance, where certain one-bit stroke encodings are introduced to represent the local information and a few paired data is used to provide the global spatial structure information of Chinese characters. Distinguished from previous studies exploiting these components of Chinese characters such as strokes through struggling to increase the model complexity, this paper provides a simple and effective way to exploit these local and global structure information as supervision for the GAN models, while almost does not increase any additional model complexity.

    \item Extensive experiments are conducted to show the effectiveness of the proposed model. Numerical results demonstrate that the mode collapse can be effectively alleviated by the use of stroke encodings and that the performance can be further improved by the suggested few-shot semi-supervised scheme, as shown in Figure \ref{Fig:mode collapse and stroke missing fig}. These provide positive answers to the questions (Q1) and (Q2) raised before. The effectiveness of the proposed model is also demonstrated through comparing with the state-of-the-art models over fourteen font generation tasks in terms of four important evaluation metrics as well as the quality of generated characters. Some generated characters yielded by the proposed model can be found in Figure \ref{Fig:results-presentation}. It can be observed that the proposed model can generate very realistic characters.

    \item We also adapt the proposed idea to some baselines based on the unpaired data to further improve their performance. Numerical results show that the proposed idea is very effective for this kind of models. The effectiveness of the proposed model for some zero-shot traditional Chinese font generation tasks is also evaluated.
\end{enumerate}

This paper is an extended version of the conference paper \cite{zeng2021strokegan}. We present the major extensions of this paper in comparison to its conference version as follows.
\begin{enumerate}
    \item[$\bullet$] In the perspective of methodology, built upon the idea of the use of stroke encodings in \cite{zeng2021strokegan} to alleviate the mode collapse, we also introduce a few-shot semi-supervised scheme (i.e., using a few paired samples as semi-supervised information) to further provide the global spatial structure information of Chinese characters. By incorporating the few-shot semi-supervised scheme, the performance of StrokeGAN suggested in the conference paper \cite{zeng2021strokegan} can be remarkably improved.

    \item[$\bullet$] In the perspective of numerical study, more generation tasks, more baselines and more evaluation metrics are considered in the numerical experiments of this paper to fully investigate the performance of the proposed model, as compared to the conference version \cite{zeng2021strokegan}. Numerical results show that the refined model StrokeGAN+ suggested in this paper significantly outperforms StrokeGAN proposed in \cite{zeng2021strokegan}. We also extend the proposed idea to other baselines and show the generalization performance of the proposed model for some zero-shot traditional Chinese font generation tasks.
\end{enumerate}

These show clearly that the extended contributions achieved by this paper are substantial in comparison to the conference paper \cite{zeng2021strokegan}.

The rest of this paper is organized as follows. Section \ref{sc:related work} provides the related work of this paper. Section \ref{sc:model} describes the proposed model in detail. We conduct extensive experiments in Section \ref{sc:experiments} to demonstrate the effectiveness of the proposed model. The generalizability and extension of the proposed idea are presented in Section \ref{sec:generalization-extension}. We conclude this paper in Section \ref{sc:conclusion}.

\section{Related Work}
\label{sc:related work}

As certain a type of artificial images, the components of Chinese characters such as strokes, radicals and skeletons are closely related to the font styles and structures of Chinese characters. Thus, incorporating these components into the generation of Chinese fonts has attracted an amount of attention in the past decade \cite{lian2012automatic,liu2012automatic,lin2015complete,lin2019font,wen2021handwritten,yuan2022se,jiang2019scfont,gao2020gan,qin2022disentangled}. In the early stage, the Chinese font generation models are mainly based on the handcrafted explicit features such as strokes and radicals \cite{lian2012automatic,liu2012automatic,lin2015complete}. Their main ideas are decomposition-and-then-assembling, that is, firstly decomposing Chinese characters into several basic components, and then assembling these components to yield new characters through some effective machine learning approaches. Although these early models are very intuitive and have good interpretability, the extraction of these handcrafted features is generally labour-intensive and expensive.

With the development of deep learning, some components of Chinese characters such as strokes, radicals and skeletons have been usually extracted by some deep neural networks and incorporated into the GAN model as certain important supervision information \cite{lin2019font,wen2021handwritten,yuan2022se,jiang2019scfont,gao2020gan,qin2022disentangled}. In \cite{lin2019font}, the authors first divided Chinese characters into strokes by adopting certain a coherent point drift algorithm and then generated new font strokes by fusing the styles of two existing font strokes and further yielded new fonts by assembling them. In \cite{wen2021handwritten}, the authors proposed a stroke refinement branch to particularly handle the generation of thin strokes involved in the Chinese font generation. In \cite{yuan2022se}, the authors exploited the skeleton information for the generation of brush handwriting Chinese font. Besides the use of an individual kind of component information, there is some literature using multi-components of Chinese characters to enhance the generation performance. In \cite{jiang2019scfont}, the authors utilized both strokes and writing trajectories for the generation of Chinese fonts, where both strokes and writing trajectories were realized by some deep neural networks. In \cite{gao2020gan}, the authors proposed a three-stage GAN model including the skeleton extraction, skeleton transformation and stroke rendering for multi-chirography image translation. In \cite{qin2022disentangled}, the authors proposed a font fusion network based on GAN and disentangled representation learning to generate brand new fonts, where the disentangled representation learning was used to obtain the stoke style and skeleton shape.

Different from previous studies struggling to increase the complexity of the model to exploit these components of Chinese characters, we aim to perform much simpler and more effective supervision for a relatively simple GAN model to achieve its full power. As shown before, the one-bit stroke encoding introduced in this paper to represent the stroke information and the use of few-shot paired samples to represent the global structure information are very simple but effective to provide supervision information for GAN models to alleviate the mode collapse issue and enhance their performance, while do not add additional model complexity. Moreover, our idea can be easily adapted to many existing GAN models to improve their performance as shown in the later experiments and the proposed model has good generalization performance in the zero-shot setting.

\begin{figure*}[!t]
\centering
\includegraphics[width=7in]{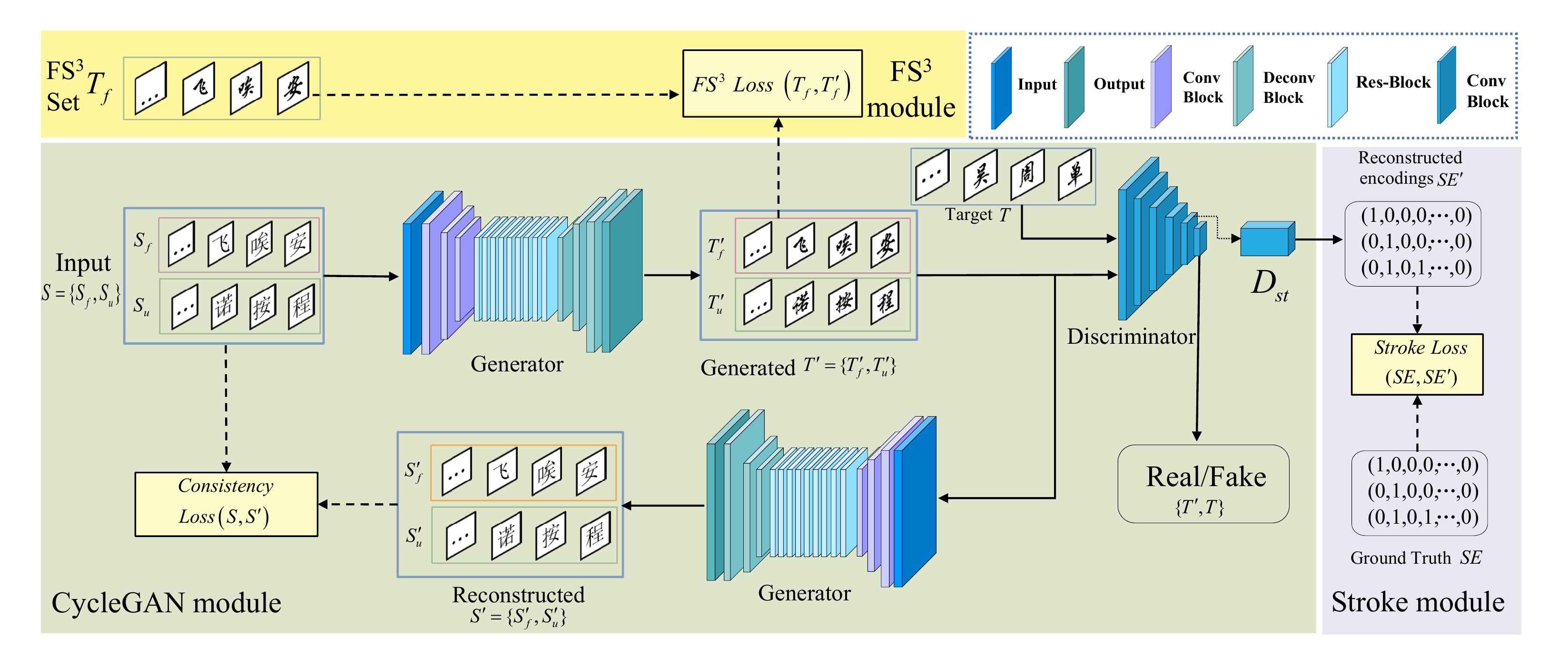} \DeclareGraphicsExtensions.
\caption{Overall architecture of StrokeGAN+. It integrates the one-bit stroke encoding and few-shot semi-supervised scheme into the well-known CycleGAN to to capture the local and global structure information respectively.
}
\label{Fig:model}
\end{figure*}

\section{Proposed Model}
\label{sc:model}

In this section, we describe the proposed model in detail. We introduce the one-bit stroke encoding as the certain embedding of stroke information and the few-shot semi-supervised training scheme in Section \ref{sc:stroke-Low-shot semi-supervised} and then describe the architecture as well as the training loss of the proposed model in Section \ref{sc:architecture}.

\subsection{Stroke Encoding and Few-shot Semi-supervised Scheme}
\label{sc:stroke-Low-shot semi-supervised}
It has been pointed out that the CycleGAN model \cite{chang2018generating} usually suffers from the mode collapse issue mainly due to the lack of effective supervision information. Inspired by the observation that strokes as the basic components of Chinese characters can be used as certain supervision information, we incorporate the stroke information into CycleGAN to alleviate the mode collapse issue. As shown in Figure \ref{Fig:stroke}(a), each Chinese character is made up of these 32 kinds of basic strokes. Based on this, we introduce a one-bit stroke encoding with 32 dimensions to represent the stroke information. We at first number these 32 types of basic strokes from 1 to 32, then for a given Chinese character, the $i$-th component of the 32-dimensional stroke encoding is set to be 1 if the $i$-th type of strokes is in the stroke set of this character and 0 otherwise. Some examples are shown in Figure \ref{Fig:stroke}(b).

However, there are some Chinese characters sharing the same one-bit stroke encodings as shown in Figure \ref{Fig:stroke}(b). This shows that such local structure information of Chinese characters may be not sufficient to supervise the model to resist the mode collapse, as shown in the fourth row of Figure \ref{Fig:mode collapse and stroke missing fig}. It can be observed from Figure \ref{Fig:mode collapse and stroke missing fig} that there are some redundant strokes in the generated characters of CycleGAN incorporated only with the one-bit stroke encoding (i.e., StrokeGAN \cite{zeng2021strokegan}). To address this issue, we suggest a few-shot semi-supervised scheme, i.e., using a few paired samples called \textit{few-shot semi-supervised} samples as additional supervision information to capture more global structure information of Chinese characters, mainly motivated by the observation that the global structure information of Chinese characters can be represented by their counterparts in the target font domain, as shown in Figure \ref{Fig:stroke}(c).

\subsection{Model Architecture and Training Loss}
\label{sc:architecture}
The proposed model is a variant of CycleGAN (dubbed \textit{StrokeGAN+}), by incorporating the stroke encoding and the few-shot semi-supervised scheme into CycleGAN. The overall architecture of the proposed model is depicted in Figure \ref{Fig:model}. The proposed model consists of three modules, i.e., the CycleGAN, stroke and few-shot semi-supervised (FS$^3$) modules. In the CycleGAN module, we use the similar architecture of CycleGAN suggested in \cite{zhu2017unpaired}, besides one more convolutional layer following the discriminator as the decoder of stroke encodings. The detailed information on networks can be found in \cite{zeng2021strokegan}. Different from CycleGAN \cite{zhu2017unpaired}, the discriminator of StrokeGAN+ undertakes two tasks. The first one is to distinguish whether the generated and real characters in the targe font domain are real or fake as done by the regular discriminator of GAN \cite{goodfellow2014generative}, and the second one is used as a decoder to reconstruct the stroke encodings of the generated characters, that is,
\begin{align*}
  D: x \rightarrow\left(D_{reg}(x), D_{st}(x)\right),
\end{align*}
where $D_{reg}(x)$ and $D_{st}(x)$ represent respectively the probability distributions over the source font domain and its stroke encoding for a given character $x$.
In the stroke module, we use the one-bit stroke encodings to supervise the model to capture more local patterns of Chinese characters, and in the FS$^3$ module, we exploit a few counterparts of the input Chinese characters in the target font domain to guide the model to extract more global patterns of Chinese characters.

Let $S = \{S_f, S_u\}$ be the sample set in the source font domain, where $S_f$ is the few-shot semi-supervised sample set containing a few samples in the source font domain, $S_u$ is the unpaired sample set. Similarly, let $T = \{T_f, T_u\}$ be the sample set in the target font domain, where $T_f$ is the paired sample set associated with $S_f$ (that is, there is a one-to-one correspondence between $S_f$ and $T_f$), and $T_u$ is the unpaired sample set in the target font domain without requiring a one-to-one correspondence between $S_u$ and $T_u$. Let $SE$ be the set of stroke encodings of $S$.

The specific workflow of the proposed model can be described as follows. We at first feed $S$ to the generator $G$ as the input, and yield a set of generated characters $T'=\{T_f', T_u'\}$ in the target font domain after $G$ (where $T'_f$ and $T'_u$ represent the sets of generated characters related to the few-shot semi-supervised samples in $S_f$ and the rest unpaired samples in $S_u$, respectively). Then, on one hand, we feed the generated characters $T'$ together with the real characters in the target font domain $T$ into the discriminator $D_{reg}$ to justify whether they are real or fake, and also yield the reconstructed stroke encodings $SE'$ of $T'$ by the decoder $D_{st}$, i.e., $SE' = D_{st}(T')$. On the other hand, we feed the generated characters $T'$ into the generator $G$ again to reconstruct the characters $S' = \{S_f', S_u'\}$ in the source font domain, as a cycle consistency task to supervise the model. Besides, we introduce a supervised scheme (i.e., the reconstruction of stroke encodings) and a semi-supervised scheme (i.e., the generation of few-shot semi-supervised samples in the target domain) to supervise the model as shown in the stroke and FS$^3$ modules respectively.

According to the above workflow, the training loss for the proposed model consists of the following four parts: the regular adversarial loss $\mathcal{L}_{adv}$, cycle consistency loss $\mathcal{L}_{cyc}$, stroke loss $\mathcal{L}_{stroke}$ and few-shot semi-supervised loss $\mathcal{L}_{FS^3}$, defined as follows:
\begin{align*}
 \mathcal{L}_{adv}(D_{reg}, G) &=\mathbb{E}_{y\sim T}[\log D_{reg}(x)] \\ &+\mathbb{E}_{x\sim S}[\log (1-D_{reg}(G(x)))],  \\
 \mathcal{L}_{cyc}(G) &=\mathbb{E}_{x\sim S}[\|x-G(G(x))\|_{1}],\\
 \mathcal{L}_{stroke}(D_{st}) &=\mathbb{E}_{(x,c)\sim (S,SE)}[\|D_{s t}(G(x))-c\|_{2}],\\
 \mathcal{L}_{FS^3}(G) &=\mathbb{E}_{(x,y) \sim (S_f,T_f)}[\|G(x)-y\|_{1}].
\end{align*}
Thus, the total training loss of the proposed model is
\begin{align*}
  \mathcal{L}_{\text {strokegan+ }}(D, G)
  &=  \mathcal{L}_{adv}(D_{reg}, G) + \lambda_{cyc} \mathcal{L}_{c y c}(G) \\
  &+\lambda_{stroke} \mathcal{L}_{stroke}(D_{st}) +\lambda_{FS^3}\mathcal{L}_{FS^3}(G),
\end{align*} 	
where $\lambda_{cyc}$, $\lambda_{stroke}$ and $\lambda_{FS^3}$  are some tuning hyperparameters. Based on the above defined loss $\mathcal{L}_{\text{strokegan+}}$, the discriminator $D$ attempts to maximize it while the generator $G$ tries to minimize it, shown as follows:
\begin{align*}
   \min_{G} \max_{D} \mathcal{L}_{\text{strokegan+}}(D, G).
\end{align*}

\section{Experiments}
\label{sc:experiments}
In this section, we conduct a series of experiments to demonstrate the effectiveness of the proposed model. We at first present experimental settings in Section \ref{sc:experimental setting}, then demonstrate the superiority of the proposed idea in reducing mode collapse in Section \ref{sc:model collapse}, and later verify the effectiveness of the introduced few-shot semi-supervised scheme in Section \ref{sec:FS3}, and finally show the effectiveness of the proposed model via comparing with the state-of-the-art models in Section \ref{sc:comparison with sota}. All experiments were carried out in Pytorch environment running Linux, AMD(R) Ryzen 7 3900x twelve-core processor ×24 CPU, and GeForce RTX 2080ti GPU.

\begin{table*}[!ht]
\renewcommand\arraystretch{1.2}
\begin{center}
\caption{Sizes of datasets for different Chinese character fonts.}
\begin{tabular}{c|c|c|c|c|c|c|c|c|c|c|c|c|c|c}\hline
 Font &ht   & zk  &fs    &hw   &hp   &st   &xk     &ls   &dl        &sj &cs &hc &bs &hh\\ \hline
 Size &3755 &2811 &3755  &2811 &2811 &2811  &3755 &3755  &2811     &3755 &2811 &2811 &1985 &2500 \\ \hline
\end{tabular}
\label{dataset}
\end{center}
\end{table*}

\subsection{Experimental Settings}
\label{sc:experimental setting}
In the next, we describe the experimental settings in detail.

 \begin{figure}[!t]
\begin{center}
	\center{\includegraphics[width=8.5cm]  {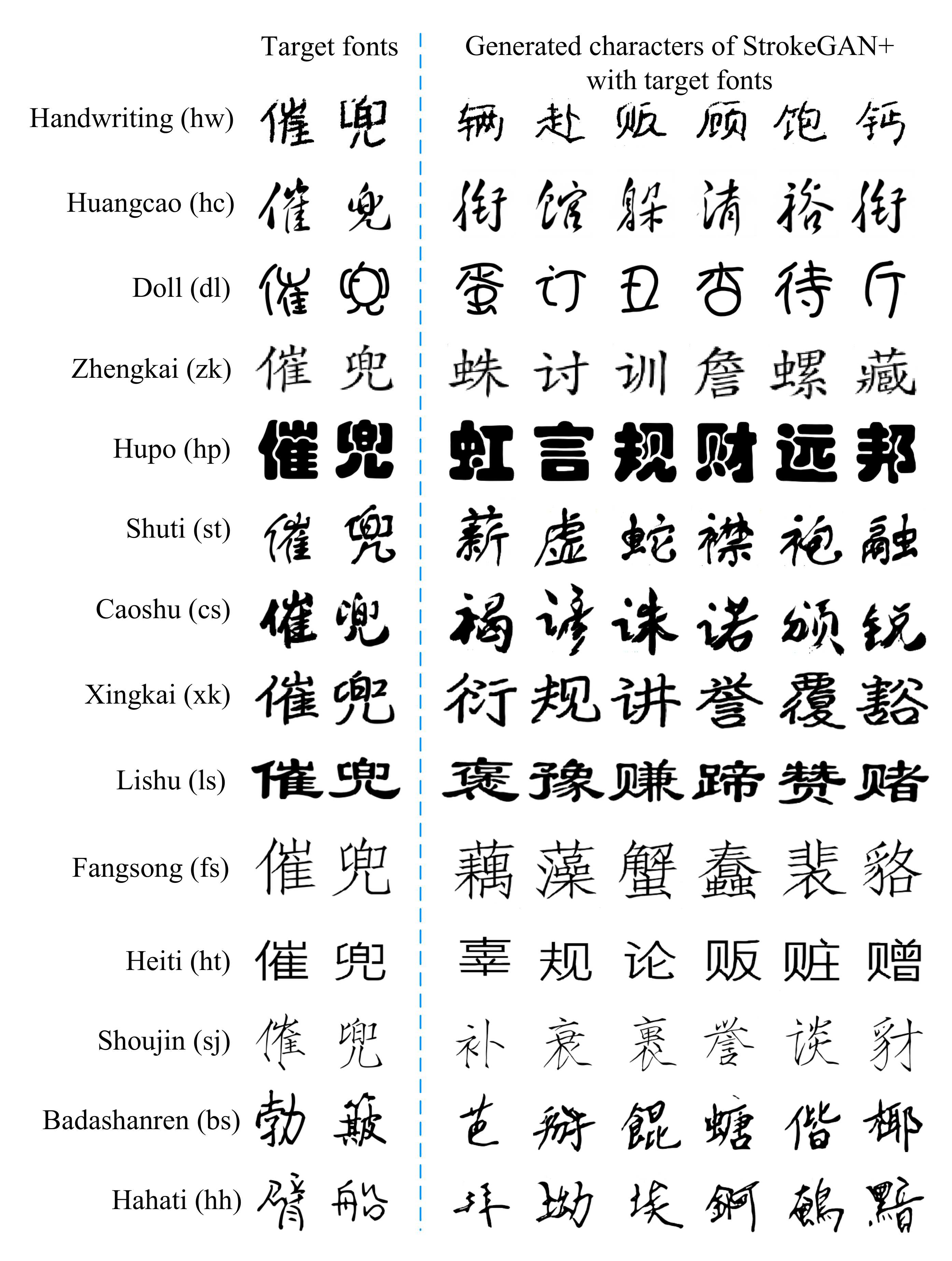}}
	\caption{Some generated characters of the proposed model.}
	\label{Fig:sample-datasets}
	\end{center}
\end{figure}

\textbf{A. Datasets.}
In our experiments, we considered fourteen Chinese fonts including three standard printing fonts \{Heiti (ht), Zhengkai (zk), Fangsong (fs)\}, the handwriting (hw) font, five pseudo-handwriting fonts \{Hupo (hp), Shuti (st), Xingkai (xk), Lishu (ls), Doll (dl)\}, and five calligraphy fonts \{Shoujin (sj), Caoshu (cs), Huangcao (hc), Badashanren (bs), Hahati (hh)\}, where the pseudo-handwriting fonts are personalized fonts designed by artistic font designers. The specific sizes of datasets are presented in Table \ref{dataset} and some samples of these datasets are presented in the left part of Figure \ref{Fig:sample-datasets}, where the handwriting dataset was collected from \textit{CASIA-HWDB1.1} and the other font datasets were collected by ourselves from the internet.
In all experiments, we used 80\% and 20\% of samples as the training and test sets, respectively.

\textbf{B. Baselines.} In this paper, we considered the following six state-of-the-art models as baselines.
\begin{itemize}
\item \textbf{Pix2Pix\cite{isola2017image}:} A typical model based on GAN and paired data.

\item \textbf{FUNIT\cite{liu2019few}:} An effective GAN model based on the disentangled representation learning and unpaired data.

\item \textbf{AttentionGAN\cite{tang2021attentiongan}:} An effective GAN model based on the attention mechanism and unpaired data.

\item \textbf{CycleGAN\cite{zhu2017unpaired}:} A typical model based on GAN and unpaired data.

\item \textbf{SQ-GAN\cite{zeng2022Square-BlockGAN}:} An improved CycleGAN model equipped with a square-block geometric transformation based self-supervised learning scheme.

\item \textbf{StrokeGAN\cite{zeng2021strokegan}:} A refined CycleGAN model incorporated with the one-bit stroke encoding for the Chinese font generation to mitigate the mode collapse.
\end{itemize}

\textbf{C. Network architectures and optimizer.}
The network architecture of StrokeGAN+ is the same as StrokeGAN \cite{zeng2021strokegan}.
The network structure of the generator of StrokeGAN+ consists of 2 convolutional layers in the down-sampling module, 9 residual modules with 2 convolutional layers of residual networks for each residual module and 2 deconvolutional layers in the up-sampling module. The network structure of the discriminator of StrokeGAN+ is similar to PatchGAN \cite{isola2017image} with 6 hidden convolutional layers and 2 convolutional layers in the output module. Moreover, the batch normalization \cite{ioffe2015batch} was used for all layers.

In our experiments, we used the popular Adam algorithm \cite{kingma2014adam} as the optimizer with the associated parameters (0.5, 0.999) in both generator and discriminator optimization subproblems. Both penalty parameters for the cycle consistency loss and stroke reconstruction loss were empirically set to be 1, and the hyperparameter $\lambda_{FS^3}$ was set by the hand-optimal way. For each generation task, we randomly selected 20\% characters as the few-shot semi-supervised samples inspired by the later experiments in Section \ref{sec:FS3}.


\textbf{D. Evaluation metrics.} We considered the following four important evaluation metrics commonly used in the literature:
\begin{itemize}
    \item \textbf{FID} (Frechet Inception Distance): FID \cite{heusel2017gans} was introduced to measure the distance between the generated and real images.

    \item \textbf{LPIPS} (Learned Perceptual Image Patch Similarity): LPIPS \cite{zhang2018unreasonable} was introduced to quantify the perceptual similarity between the generated and real images.

     \item \textbf{PSNR} (Peak Signal-to-Noise Ratio): PSNR is a commonly used metric for the assessment of image quality, defined as the ratio between the maximum possible power of an image and the power of corrupting noise that affects the quality of its representation.

    \item \textbf{SSIM} (Structural Similarity): SSIM \cite{Wang2004-SSIM} is a perceptual metric that quantifies image quality degradation caused by processing.
\end{itemize}
In general, smaller FID and LPIPS values imply better quality, while larger SSIM and PSNR values imply better quality. Both LPIPS and SSIM lie in $(0, 1)$.


\subsection{Superiority in Reducing Mode Collapse}
\label{sc:model collapse}
In this section, we conducted three generation tasks including the translations from Fangsong to Zhengkai (fs$\rightarrow$zk), from Fangsong to Heiti (fs$\rightarrow$ht), and from Fangsong to Lishu (fs$\rightarrow$ls) to show the superiority of the proposed idea in reducing mode collapse. Besides CycleGAN, we also compared the performance of the proposed model with SQ-GAN \cite{zeng2022Square-BlockGAN} and StrokeGAN \cite{zeng2021strokegan}. The comparison results are shown in Figure \ref{Fig:mode collapse and stroke missing fig} and Table \ref{Table: mode collapse}.

From Figure \ref{Fig:mode collapse and stroke missing fig}, the well-known CycleGAN model suffers from the mode collapse issue in these three generation tasks, while the mode collapse issue can be effectively addressed by the proposed idea. Specifically, as shown in the second row of Figure \ref{Fig:mode collapse and stroke missing fig}, CycleGAN produces the same patterns for all different inputs and yields the incorrect characters, while the mode collapse issue can be significantly alleviated by incorporating the one-bit stroke encoding into CycleGAN as shown in the fourth row of Figure \ref{Fig:mode collapse and stroke missing fig} and can be further reduced via the introduced few-shot semi-supervised scheme as shown in the last row of Figure \ref{Fig:mode collapse and stroke missing fig}. By the third row of Figure \ref{Fig:mode collapse and stroke missing fig}, although the mode collapse issue of CycleGAN can be alleviated to a certain extent by integrating the square-block geometric transformation based self-supervised scheme, SQ-GAN still suffers from the mode collaspe issue in the first generation task. This shows that such square-block geometric transformation based self-supervised scheme should be insufficient to thoroughly tackle the mode collapse issue. As shown in the fourth row of Figure \ref{Fig:mode collapse and stroke missing fig}, the StrokeGAN model can effectively alleviate the mode collapse issue of CycleGAN via equipping with a one-bit stroke encoding as supervision information. This shows that the introduced one-bit stroke encoding can capture more important patterns of Chinese characters during the generation. However, StrokeGAN may suffer from the issue of missing stroke or generating redundant strokes, that is, missing some strokes or generating some additional strokes in the generated characters as shown in the second and third generation tasks. This shows that such simple one-bit stroke encoding may be also not enough to supervise CycleGAN to generate high-quality characters. To address this issue, we further use the few-shot semi-supervised samples as certain auxiliary information to explore the global structure information of Chinese characters. As shown in the fifth row of Figure \ref{Fig:mode collapse and stroke missing fig}, the quality of generated Chinese characters of StrokeGAN+ is much better than that of StrokeGAN, mainly due to the use of few-shot semi-supervised samples. This demonstrates that these few-shot semi-supervised samples can provide some useful global structure information during the training procedure.

Moreover, from Table \ref{Table: mode collapse}, since the mode collapse happens for CycleGAN in these three font generation tasks, the FID values of CycleGAN are abnormally large, while these values can be remarkably reduced by equipping with the square-block geometric transformation based self-supervised scheme in SQ-GAN \cite{zeng2022Square-BlockGAN} or the one-bit stroke encoding in StrokeGAN \cite{zeng2021strokegan}. It can be observed that the FID value of SQ-GAN in the first generation task is also abnormally large mainly due to the mode collapse issue as also shown in the third row of Figure \ref{Fig:mode collapse and stroke missing fig}, while StrokeGAN as well as its refined version StrokeGAN+ suggested in this paper do not suffer from the model collapse issue in these concerned font generation tasks. From the third and fourth rows of Table \ref{Table: mode collapse}, the performance of StrokeGAN can be further improved by the use of few-shot semi-supervised samples as the global structure supervision information in terms of all these four evaluation metrics. These clearly show the effectiveness of the proposed idea.

\begin{table}[!ht]
\renewcommand\arraystretch{1.2}
\tabcolsep=0.38cm
    \caption{Superiority of the proposed model in reducing mode collapse. The best results are marked in bold.}
	\label{Table: mode collapse}
\begin{center}
\begin{tabular}{c|c|c|c|c}
\hline
Metric & Model  & fs$\rightarrow$zk  & fs$\rightarrow$ht  & fs$\rightarrow$ls \\ \hline\hline
\multirow{4}{*}{FID$\downarrow$}
& CycleGAN    & 192.52     & 304.33    & 287.81\\
& SQ-GAN      & 269.54     & 32.24     & 26.01\\
& StrokeGAN   & 32.46      & 45.98     & 56.10\\
& StrokeGAN+  & {\bf 29.83}      & {\bf 25.22}     & {\bf 33.80}\\\hline\hline
\multirow{4}{*}{LPIPS$\downarrow$}
& CycleGAN    & 0.407      & 0.384     & 0.350\\
& SQ-GAN      & 0.449      & 0.214     & 0.190\\
& StrokeGAN   & 0.194      & 0.289     & 0.21\\
& StrokeGAN+  & {\bf 0.158}      & {\bf 0.132}     & {\bf 0.150}\\\hline\hline
\multirow{4}{*}{PSNR$\uparrow$}
& CycleGAN    & 7.37       & 7.06      & 7.42\\
& SQ-GAN      & 7.12       & 7.57      & 8.82\\
& StrokeGAN   & 9.21       & 7.28      & 8.73\\
& StrokeGAN+  & {\bf 10.34}      & {\bf 9.95}      & {\bf 10.22}\\\hline\hline
\multirow{4}{*}{SSIM$\uparrow$}
& CycleGAN    & 0.575      & 0.346     & 0.455\\
& SQ-GAN      & 0.382      & 0.482     & 0.597\\
& StrokeGAN   & 0.533      & 0.446     & 0.583\\
& StrokeGAN+  & {\bf 0.664}      & {\bf 0.636}     & {\bf 0.653}\\\hline
\end{tabular}
\end{center}
\end{table}

\subsection{On Few-Shot Semi-Supervised Schemes}
\label{sec:FS3}
As shown in the last section, the few-shot semi-supervised samples are crucial for the proposed model. In this section, we suggested two strategies to determine the few-shot semi-supervised samples, i.e., a random scheme and a deterministic scheme, and compared them in some font generation tasks, and finally provided some comparisons between the suggested few-shot semi-supervised scheme and the associated copy data augmentation scheme.

{\bf A. Two Few-shot Semi-supervised Schemes.}
In this paper, we suggested two strategies for the selection of few-shot semi-supervised samples as follows.
\begin{enumerate}
    \item[(a)] \textbf{Random strategy:} The few-shot semi-supervised samples are selected randomly from the training set with certain percentages.
    \item[(b)] \textbf{Determinstic strategy:} According to \cite{wen2021handwritten}, there are 450 single characters and 300 compound Chinese characters that cover all structural information of Chinese characters. Thus, we can select the few-shot semi-supervised samples from such a set of Chinese characters of size 750 to capture more structural information of Chinese characters.
\end{enumerate}

 \begin{figure}[h]
\begin{center}
	\center{\includegraphics[width=7cm]  {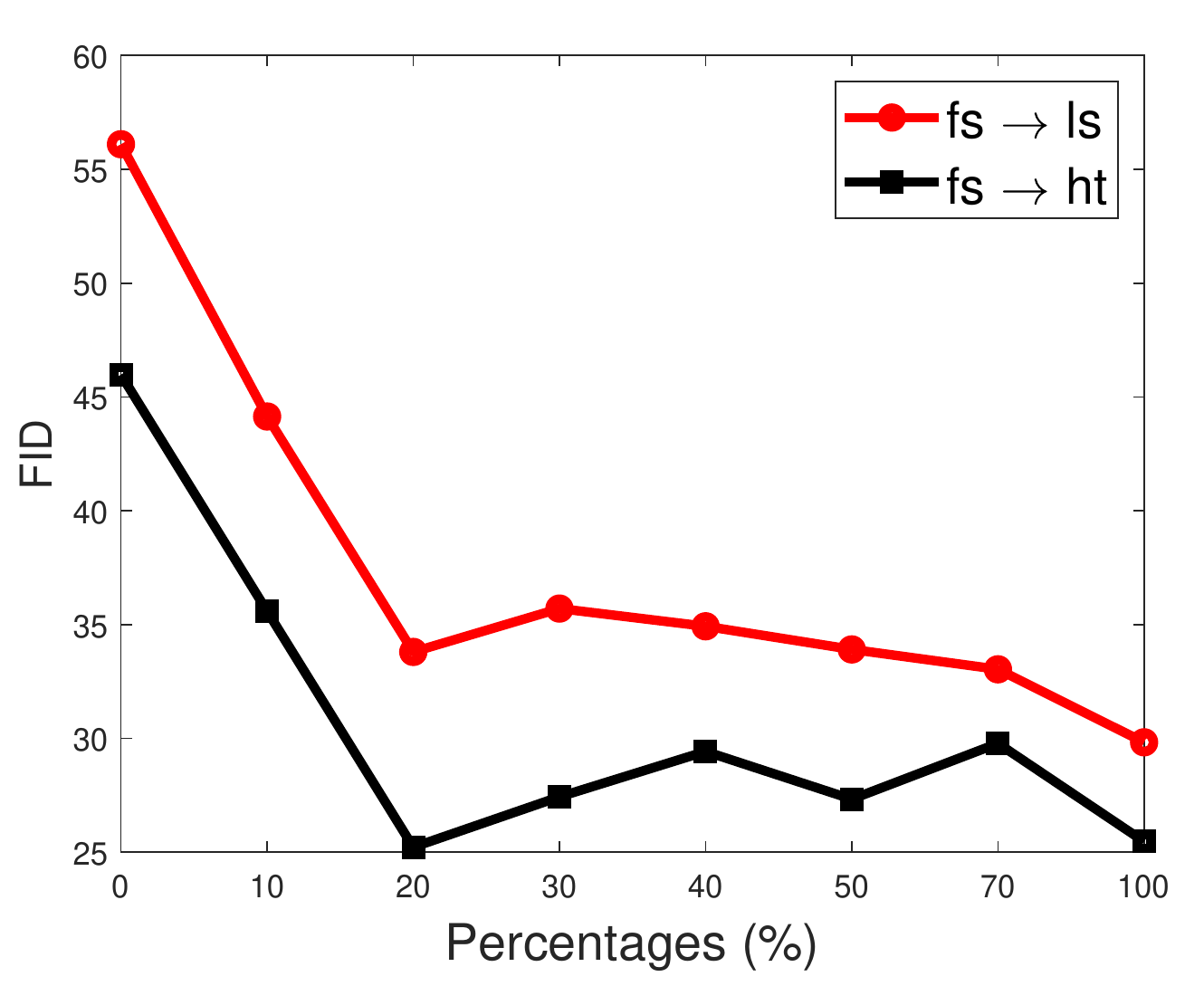}}
	\caption{Trends of FID values of StrokeGAN+ using random selection strategies with different percentages for few-shot semi-supervised samples over two generation tasks.}
	\label{Fig:FID-percentage-Low-shot semi-supervised samples}
	\end{center}
\end{figure}

\begin{table*}[t]
\renewcommand\arraystretch{1.2}
\tabcolsep=0.4cm
    \caption{Comparison on the performance of two kinds of few-shot semi-supervised schemes over two printed font generation tasks.}
\label{Tab:comp_printed_rand_determine}
\begin{center}

\begin{tabular}{c|c|c|c|c|c|c}
\hline
Metric & Task  & StrokeGAN    & StrokeGAN+$_{r20\%}$           & StrokeGAN+$_{d250}$ & StrokeGAN+$_{d500}$   & StrokeGAN+$_{d750}$            \\ \hline\hline
\multirow{2}{*}{FID$\downarrow$}   & fs$\rightarrow$ht & 45.98 & 25.22          & 30.98  & 21.08 & \textbf{19.45} \\ \cline{2-7}
                       & fs$\rightarrow$ls& 56.10 & 33.80          & 31.70  & 23.53 & \textbf{19.03} \\ \hline\hline
\multirow{2}{*}{LPIPS$\downarrow$} &fs$\rightarrow$ht & 0.289 & 0.132          & 0.141  & \textbf{0.125} & 0.131 \\ \cline{2-7}
                       & fs$\rightarrow$ls & 0.215 & 0.150          & 0.138  & 0.135 & \textbf{0.134} \\ \hline\hline
\multirow{2}{*}{PSNR$\uparrow$}  & fs$\rightarrow$ht & 7.28  & \textbf{9.95}  & 9.27   & 9.77  & 9.59           \\ \cline{2-7}
                       & fs$\rightarrow$ls & 8.73  & 10.22          & 10.18  & 10.27 & \textbf{10.31} \\ \hline\hline
\multirow{2}{*}{SSIM$\uparrow$}  & fs$\rightarrow$ht & 0.446 & \textbf{0.636} & 0.593  & 0.624 & 0.617          \\ \cline{2-7}
                       & fs$\rightarrow$ls & 0.583 & 0.653          & 0.649  & 0.652 & \textbf{0.656} \\ \hline
\end{tabular}
\end{center}
\end{table*}

\begin{table*}[!h]
\renewcommand\arraystretch{1.2}
\begin{center}
\tabcolsep=0.12cm

\caption{Comparison on the performance of two kinds of few-shot semi-supervised schemes over two calligraphy font generation tasks.}
\label{Tab:comp_calligraphy_rand_determine}
\begin{tabular}{c|cccc|cccc}
\hline
Task  & \multicolumn{4}{c|}{fs$\rightarrow$bs}                             & \multicolumn{4}{c}{fs$\rightarrow$hh}                               \\ \hline
Model     &{\scriptsize StrokeGAN} & {\scriptsize StrokeGAN+$_{d490}$} & {\scriptsize StrokeGAN+$_{r490}$}   &{\scriptsize StrokeGAN+$_{r20\%}$}  & {\scriptsize StrokeGAN} & {\scriptsize StrokeGAN+$_{d600}$} & {\scriptsize StrokeGAN+$_{r600}$}   &{\scriptsize StrokeGAN+$_{r20\%}$} \\ \hline
FID$\downarrow$   & 66.39   & \textbf{49.57} & 53.34          & 52.90        & 64.24     & 48.51         & \textbf{46.48} & 48.05          \\ \hline
LPIPS$\downarrow$ & 0.278   & 0.241          & \textbf{0.240} & 0.246        & 0.309     & 0.314         & \textbf{0.292} & 0.294          \\ \hline
PSNR$\uparrow$  & 6.82    & 7.45           & \textbf{7.46}  & 7.44         & 7.38      & \textbf{7.58} & 7.52           & 7.49           \\ \hline
SSIM$\uparrow$  & 0.445   & \textbf{0.491} & 0.490          & 0.489        & 0.484     & 0.483         & \textbf{0.493} & 0.490          \\ \hline

\end{tabular}
\end{center}
\end{table*}

{\bf B. Comparison on Few-shot Semi-supervised Schemes.}
In the following, we conducted a series of experiments to compare the suggested two kinds of few-shot semi-supervised schemes.

In the first experiment, we considered the performance of random strategies with different percentages over two generation tasks, i.e., \{fs$\rightarrow$ls, fs$\rightarrow$ht\}. For each generation task, we considered eight different percentages of paired samples as the few-shot semi-supervised samples, i.e., \{0\%, 10\%, 20\%, 30\%, 40\%, 50\%, 70\%, 100\%\}, where StrokeGAN+ with 0\% paired samples reduces to the StrokeGAN model \cite{zeng2021strokegan}. The trends of FID values of StrokeGAN+ with different percentages of paired samples over two generation tasks are depicted in Figure \ref{Fig:FID-percentage-Low-shot semi-supervised samples}. From Figure \ref{Fig:FID-percentage-Low-shot semi-supervised samples}, the FID values of StrokeGAN+ dramatically decrease with the percentage of paired samples increasing to 20\%, and become stable when the percentage of the used paired samples as the few-shot semi-supervised samples are larger than 20\%. These show that on one hand, using the few-shot semi-supervised samples as the supervision information is very effective for the Chinese font generation as depicted in Figure \ref{Fig:FID-percentage-Low-shot semi-supervised samples}, where the FID values of StrokeGAN+ with 10\% paired samples as the few-shot semi-supervised samples are much better than that of StrokeGAN (i.e., StrokeGAN+ without few-shot semi-supervised samples); and on the other hand, only a few paired samples (say, 20\% paired training data including 600 characters in these experiments) as the few-shot semi-supervised samples are in general adequate to provide the supervision for the Chinese font generation models. These show the feasibility of our idea of using few-shot semi-supervised samples as the global structure supervision information for the Chinese font generation models. In the later experiments, we used 20\% paired training data as the default random few-shot semi-supervised scheme.

Besides these eight random cases, we also considered three deterministic selection schemes, i.e., using 250, 500 and 750 characters of the special set of Chinese characters determined in advance according to the literature \cite{wen2021handwritten} as the few-shot semi-supervised samples. The proposed StrokeGAN+ models with the 20\% random scheme (randomly selecting 600 paired characters as the few-shot semi-supervised samples), and 250, 500 and 750 deterministic schemes are denoted as StrokeGAN+$_{r20\%}$, StrokeGAN+$_{d250}$, StrokeGAN+$_{d500}$, and StrokeGAN+$_{d750}$, respectively. The comparison results among these schemes over the two font generation tasks \{fs$\rightarrow$ls, fs$\rightarrow$ht\} are presented in Table \ref{Tab:comp_printed_rand_determine}. From Table \ref{Tab:comp_printed_rand_determine}, the proposed StrokeGAN+ models with all few-shot semi-supervised schemes substantially improve the performance of StrokeGAN in terms of the concerned four evaluation metrics. When concerning the performance among these few-shot semi-supervised schemes, the performance of StrokeGAN+ with the deterministic strategy using a total of 750 characters is generally the best over the considered two generation tasks, and the performance of the deterministic strategy with 500 characters is slightly better than that of the random strategy with 20\% training data (i.e., 600 characters in this experiment), which is slightly better than the performance of the deterministic strategy with 250 characters. Such phenomenon is mainly because more structure information of Chinese characters can be captured by choosing the few-shot semi-supervised samples from these 750 representative characters.

Although the deterministic few-shot semi-supervised scheme can generally achieve better performance than the associated random scheme with the same size of few-shot semi-supervised samples, it might be unrealistic when applied to the generation tasks of some calligraphy fonts due to the limited samples for these calligraphy fonts. In the following, we considered the performance of these two kinds of few-shot semi-supervised schemes over two calligraphy font generation tasks, i.e., from Fangsong font to the Badashanren font (fs$\rightarrow$bs) and from Fangsong font to the Hahati font (fs$\rightarrow$hh). The sample sets of Badashanren and Hahati fonts for training only contain 490 and 600 characters in the specified 750 character set, respectively. For these two tasks, we considered two random schemes with the different numbers of characters, i.e., the total number of representative characters in the specified Chinese character set (490 for Badashanren font and 600 for Hahati font), and 20\% training samples (317 for Badashanren font and 400 for Hahati font). The proposed models StrokeGAN+ with 490 deterministic scheme, 490 random scheme and 20\% (including 317 characters) random scheme for the generation task \{fs$\rightarrow$bs\} are denoted as StrokeGAN+$_{d490}$, StrokeGAN+$_{r490}$ and StrokeGAN+$_{r20\%}$, respectively. The associated models for the generation task \{fs$\rightarrow$hh\} are denoted similarly. The comparison results among the proposed StrokeGAN+ models with these two kinds of few-shot semi-supervised schemes as well as StrokeGAN are presented in Table \ref{Tab:comp_calligraphy_rand_determine}.

From Table \ref{Tab:comp_calligraphy_rand_determine}, all the proposed StrokeGAN+ models outperform the StrokeGAN model with substantial improvements in terms of four important evaluation metrics over the concerned two calligraphy font generation tasks, through integrating the few-shot semi-supervised schemes. When concerning the performance of these two kinds of few-shot semi-supervised schemes over the font generation task \{fs$\rightarrow$bs\}, the deterministic scheme with 490 characters from the specified character set achieves better performance in terms of FID and SSIM and slightly worse performance in terms of LPIPS and PSNR as compared to the random scheme with 490 characters randomly selected from the training set, while the performance of the random scheme with 20\% training samples (317 characters included in this case) is comparable to that the deterministic scheme with 490 characters. A similar claim can be also concluded over another calligraphy font generation task, i.e., \{fs$\rightarrow$hh\}. These show that we can use the 20\% random scheme as the default scheme in the proposed model for the sake of convenience.

\textbf{C. Few-shot Semi-supervised Scheme Vs. Data Augmentation.}
Since an additional few-shot semi-supervised sample set is used in StrokeGAN+, a natural question that arises here is whether the improvement achieved by the use of few-shot semi-supervised samples is only brought by the associated copy data augmentation. To answer this question, we implemented several experiments over four generation tasks \{Zhengkai$\rightarrow$Hupo, Fangsong$\rightarrow$Heiti, Fangsong$\rightarrow$Lishu, Zhengkai$\rightarrow$Shuti\} and compared the performance of StrokeGAN+ (using 20\% random few-shot semi-supervised scheme) with that of StrokeGAN models equipped with different data augmentation schemes,  i.e.,
\begin{itemize}
    \item \text{StrokeGAN}$_{0\%}$: the original StrokeGAN without using paired samples;
    \item \text{StrokeGAN}$_{20\%}$: StrokeGAN with 20\% paired samples as unpaired data;
     \item \text{StrokeGAN}$_{100\%}$: StrokeGAN with 100\% paired samples as unpaired data, that is, simply doubling the datasets for training.
\end{itemize}

 \begin{figure*}[!ht]
\begin{center}
	\center{\includegraphics[width=18cm]  {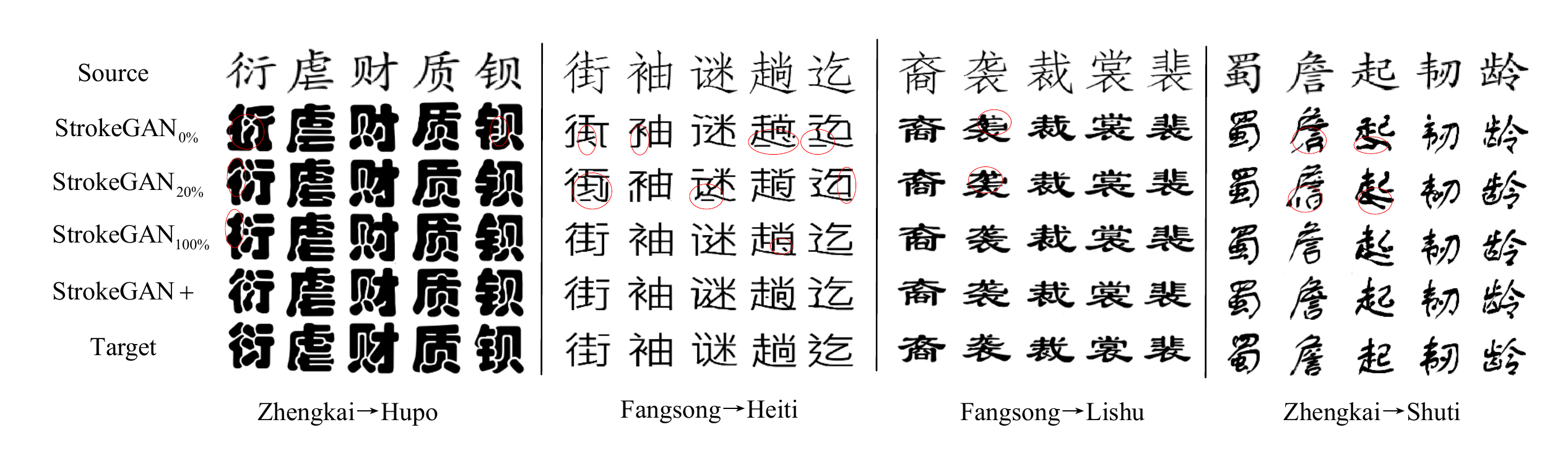}}
	\caption{Comparisons on visualization results of StrokeGAN with the few-shot semi-supervised scheme and the simple copy data augmentation scheme over four generation tasks.}
	\label{Fig:Low-shot vs. augmentation-visual}
	\end{center}
\end{figure*}

\begin{table}[!ht]
\renewcommand\arraystretch{1.2}
\caption{Quantitative comparisons on performance of StrokeGAN with the few-shot semi-supervised scheme and the simple copy data augmentation scheme over four generation tasks.}
 \label{Table:Low-shot vs data augmentation}

\begin{tabular}{c|c|cccc}
\hline                  & Model               & zk$\rightarrow$hp & fs$\rightarrow$ht & fs$\rightarrow$ls & zk$\rightarrow$st \\ \hline\hline
\multirow{4}{*}{FID$\downarrow$}   & {StrokeGAN}$_{0\%}$           & 78.67          & 45.98          & 56.10          &85.20          \\
                        & StrokeGAN$_{20\%}$  & 74.71          & 52.88          & 47.94          & 69.77          \\
                        & StrokeGAN$_{100\%}$ & 61.84          &  36.26               & 44.91          & 60.02          \\
                        & StrokeGAN+     & \textbf{44.72} & \textbf{25.22} & \textbf{33.80} & \textbf{52.33} \\ \hline\hline
\multirow{4}{*}{LPIPS$\downarrow$} & {StrokeGAN}$_{0\%}$          & 0.324          & 0.289          & 0.215          &0.308          \\
                        & StrokeGAN$_{20\%}$  & 0.324          & 0.253          & 0.191          & 0.396          \\
                        & StrokeGAN$_{100\%}$ & 0.275          &     0.246            & 0.187          & 0.258          \\
                        & StrokeGAN+     &\textbf{ 0.257 }         & \textbf{0.132} & \textbf{0.150} & \textbf{0.225} \\ \hline\hline
\multirow{4}{*}{PSNR$\uparrow$}  & {StrokeGAN}$_{0\%}$           & 6.10           & 7.28           & 8.73           &6.83           \\
                        & StrokeGAN$_{20\%}$ & 6.10           & 7.00           & 8.76           & 7.36           \\
                        & StrokeGAN$_{100\%}$ & 7.03           &       7.34          & 8.74           & 7.10           \\
                        & StrokeGAN+     & \textbf{7.16}  & \textbf{9.95}  & \textbf{10.22} & \textbf{7.78}  \\ \hline\hline
\multirow{4}{*}{SSIM$\uparrow$}  & {StrokeGAN}$_{0\%}$           & 0.321          & 0.446          & 0.583          & 0.468          \\
                        & StrokeGAN$_{20\%}$  & 0.322          & 0.429          & 0.591          & 0.501          \\
                        & StrokeGAN$_{100\%}$ & 0.383          &   0.433              & 0.589          & 0.486          \\
                        & StrokeGAN+      & \textbf{0.406 }         & \textbf{0.636} & \textbf{0.653} & \textbf{0.520} \\ \hline
\end{tabular}
\end{table}

Some quantitative comparison results are presented in Table \ref{Table:Low-shot vs data augmentation}. As shown in the second and third rows of Table \ref{Table:Low-shot vs data augmentation}, the improvement yielded by the data augmentation with only 20\% additional unpaired data is relatively limited, while StrokeGAN+ with the same 20\% few-shot semi-supervised samples significantly outperforms StrokeGAN$_{20\%}$, and even much better than StrokeGAN$_{100\%}$ using the double size of datasets. We also present some comparisons of visualization results of these concerned models in Figure \ref{Fig:Low-shot vs. augmentation-visual}. It can be observed from Figure \ref{Fig:Low-shot vs. augmentation-visual} that the proposed StrokeGAN+ with only 20\% random few-shot semi-supervised scheme outperforms that of StrokeGAN models equipped with different copy data augmentation schemes. These show clearly that the improvement achieved by the use of few-shot semi-supervised scheme is mainly because the few-shot semi-supervised samples used can capture certain global structures of Chinese characters.

\begin{table*}[!t]
\renewcommand\arraystretch{1.2}
    \caption{Comparision on the performance of the proposed model and state-of-the-art models in nine printing or handwriting font generation tasks. The best and second best results are marked in bold and blue color respectively.}
\label{Tab:comparison-SOTA-printed}
\begin{center}
\begin{tabular}{c|c|c|c|c|c|c|c|c|c|c}
\hline
& Model  & fs$\rightarrow$xk    & zk$\rightarrow$hp & zk$\rightarrow$hw & fs$\rightarrow$ht & fs$\rightarrow$zk & zk$\rightarrow$st & zk$\rightarrow$dl & fs$\rightarrow$ls   & zk$\rightarrow$fs \\ \hline\hline
\multirow{7}{*}{\rotatebox{90}{FID$\downarrow$}}
& Pix2Pix  & 53.95       & 195.01    & 158.59  & 118.25  & 60.46   & 184.86 & 139.09  & 44.75   & 91.72  \\
& FUNIT    & 56.24       & 89.12     & 116.95   & 142.13  & 155.86  & \textbf{44.11}  &  79.50  & 67.77     & 86.91\\
& AttentionGAN  & \color{blue}{21.60}  & 58.80 & \textbf{37.95}  &\textcolor{blue}{ 28.31}  & 52.68     & 87.17  & 64.47      & \textbf{14.95}    & \textcolor{blue}{33.12}\\
& CycleGAN      & 36.48       & 71.44 & 66.61  & 304.33 & 192.53  & 79.33 & 64.58  & 287.82  & 32.325\\
& SQ-GAN        & 49.40    &\color{blue}{56.75}   & 59.88  & 32.24  & 269.54  & 71.69& 65.76  & \textcolor{blue}{26.01}    & 36.64\\
& StrokeGAN     & 24.33     & 78.67  & 56.46  & 45.98  & \textcolor{blue}{32.46}  & 85.20 & \textcolor{blue}{61.61}  & 56.10    & 33.86\\
&StrokeGAN+     & \textbf{18.25}      & \textbf{44.72}   & \color{blue}{42.83}  & \textbf{25.22}  & \textbf{29.83}  & \textcolor{blue}{52.33}  & \textbf{38.37}  & 33.80   & \textbf{25.49}\\\hline\hline
\multirow{7}{*}{\rotatebox{90}{LPIPS$\downarrow$}}
& Pix2Pix   & \color{blue}0.184       & 0.292    & 0.340  & 0.179  & \textcolor{blue}{0.175}  & 0.353  & \textcolor{blue}{0.331}  & 0.162    & 0.167\\
& FUNIT    & 0.269     & 0.335   & 0.417   & 0.335  & 0.327  & 0.335  & 0.350  & 0.303    & 0.231\\
& AttentionGAN  & 0.209     & \color{blue} 0.276  & \color{blue} 0.327  & \textcolor{blue}{0.162} & 0.239  & 0.303  & 0.370  & \textbf{0.142}    & 0.141\\
& CycleGAN  &0.216    &0.328   & 0.407  & 0.384  & 0.408  & \textcolor{blue}{0.283}  & 0.351  & 0.350    & \textcolor{blue}{0.131}\\
& SQ-GAN  & 0.219      & 0.297  & \textbf{0.320}    & 0.214  & 0.449  & 0.301  & 0.365  & 0.190   & 0.144\\
& StrokeGAN  & 0.234     & 0.324  & 0.338 & 0.289  & 0.194  & 0.308  & 0.354  & 0.215   &  0.161\\
& StrokeGAN+  & \textbf{0.160}        & \textbf{0.256}    & 0.331  & \textbf{0.132} & \textbf{0.158}   & \textbf{0.225} & \textbf{0.308}  & \textcolor{blue}{0.150}   & \textbf{0.128}\\ \hline\hline
\multirow{7}{*}{\rotatebox{90}{PSNR$\uparrow$}}
& Pix2Pix  & \color{blue}8.63     & 6.39   & 5.66   & \textcolor{blue}{9.00} & \textbf{10.51}   & 6.61  & \textcolor{blue}{5.49}  & \textcolor{blue}{10.28}    & 9.53\\
& FUNIT   & 7.30    & \textcolor{blue}{6.95}   & 5.45  & 6.70  & 8.87  & 6.49  & 5.14  & 7.22     & 9.13\\
& AttentionGAN  & 8.17   & 6.76  & 5.71  & 8.90  & 9.18  & 6.91  & 5.17  & \textbf{10.51}    & \textcolor{blue}{9.80}\\
& CycleGAN  & 7.82       & 6.77   & 5.40   & 7.06   & 7.37  &\textcolor{blue}{7.26}   & 5.17   & 7.47   & 9.67\\
& SQ-GAN  & 7.95      & 6.68   & \color{blue}5.79   & 7.57   & 7.12  & 7.06  & 5.05  & 8.82   & 9.48\\
& StrokeGAN  & 7.61    &6.10   & 5.53   & 7.28  & 9.21  &6.83  & 5.11  & 8.73   & 9.03\\
&StrokeGAN+   & \textbf{8.89}     & \textbf{7.16}  &\textbf{5.99}   & \textbf{9.95}  & \textcolor{blue}{10.34}  & \textbf{7.78}  & \textbf{5.67}  & 10.22    & \textbf{10.25}\\ \hline\hline
 \multirow{7}{*}{\rotatebox{90}{SSIM$\uparrow$}}
& Pix2Pix  & \color{blue} 0.545     & 0.367  & 0.305  & 0.548  & \textcolor{blue}{0.582}  & 0.442  & \textcolor{blue}{0.296}  & \textcolor{blue}{0.640}   & 0.553\\
& FUNIT  & 0.449     & \textbf{0.472}  & 0.305   & 0.350  & 0.511  & 0.417  & 0.262  & 0.448    & 0.521\\
& AttentionGAN   & 0.527    & 0.351   & 0.335   & \textcolor{blue}{0.564}  & 0.515  & 0.457   & 0.277  & \textbf{0.660}    & \textcolor{blue}{0.578}\\
& CycleGAN  & 0.510     & 0.376  & 0.324  & 0.346  & 0.575  &\textcolor{blue}{0.495} & 0.273  & 0.455   & 0.576\\
& SQ-GAN   & 0.516     & 0.363  & \color{blue} 0.338   & 0.482  & 0.382 & 0.448 & 0.263  & 0.597   & 0.567\\
& StrokeGAN  & 0.502    & 0.321  & 0.318  & 0.446  & 0.533   & 0.468  & 0.264  & 0.583    & 0.534\\
& StrokeGAN+   & \textbf{0.572}      & \color{blue} 0.406   & \textbf{0.363}  & \textbf{0.636}  &  \textbf{0.665}  & \textbf{0.520}   & \textbf{0.317}  & 0.653   & \textbf{0.618}\\ \hline\hline
\end{tabular}
\end{center}
\end{table*}

\begin{table}[h]
\renewcommand\arraystretch{1.2}
\tabcolsep=0.19cm
\caption{Comparison with state-of-the-art methods in five calligraphy font generation tasks. The best and second best results are marked in bold and blue color respectively.}
\label{Tab:comparison-SOTA-calligraphy}
\begin{tabular}{c|c|c|c|c|c|c}
\hline\hline
                        & Model             & fs$\rightarrow$bs                        & fs$\rightarrow$hh                        & zk$\rightarrow$cs                        & xk$\rightarrow$sj                        & ls$\rightarrow$hc                               \\ \hline \hline
         \multirow{7}{*}{\rotatebox{90}{FID$\downarrow$}}                & Pix2Pix      & 102.26                       & 104.55                       & 124.18                       & 74.09                        & 103.15            \\
                        & FUNIT        & 61.75                        & 78.93                        & 53.75                        & 51.41                        & 94.10           \\
                        & AttentionGAN & 56.51                        & 92.15                        & \textbf{40.23}               & 70.23                        & {\color{blue} 28.66}      \\
                        & CycleGAN     & 75.46                        & 77.08                        & 80.41                        & 47.35                        & 36.05     \\
                        & SQ-GAN       & {\color{blue} 55.55} & {\color{blue} 59.32} & 58.48                        & 36.50                        & 45.27      \\
                        & StrokeGAN    & 66.39                        & 64.24                        & 62.37                        & {\color{blue} 36.42} & 43.58     \\
   & StrokeGAN+   & \textbf{52.90}               & \textbf{48.05}               & {\color{blue} 49.85} & \textbf{31.27}               & \textbf{21.10}                        \\ \hline \hline
         \multirow{7}{*}{\rotatebox{90}{LPIPS$\downarrow$}}                & Pix2Pix      & 0.269                        & 0.317                        & {\color{blue}0.303 }                       & 0.224                        & 0.352   \\
                        & FUNIT        & 0.301                        & 0.343                        & 0.339                        & 0.295                        & 0.431     \\
                        & AttentionGAN & 0.370                        & 0.330                        &  0.306 & 0.225                        & {\color{blue} 0.315}         \\
                        & CycleGAN     & {\color{blue} 0.256} & 0.327                        & 0.316                        & {\color{blue} 0.211} & 0.321 \\
                        & SQ-GAN       & 0.290                        & 0.323                        & 0.319                        & 0.250                        & 0.329  \\
                        & StrokeGAN    & 0.278                        & {\color{blue} 0.309} & 0.331                        & 0.267                        & 0.316   \\
& StrokeGAN+   & \textbf{0.246}               & \textbf{0.294}               & \textbf{0.280}               & \textbf{0.207}               & \textbf{0.221}                        \\ \hline \hline
   \multirow{7}{*}{\rotatebox{90}{PSNR$\uparrow$}}                     & Pix2Pix      & \textbf{7.59}                & \textbf{7.56}                & {\color{blue}7.51}                         & \textbf{9.91}                & 7.57
                      \\  & FUNIT        & 7.27                         & 7.02                         & 6.05                         & 9.47                         & \textbf{8.27}
                     \\   & AttentionGAN & 7.41                         & 7.41                         & 7.35                         & 9.60                         & 7.72
                      \\  & CycleGAN     & 7.13                         & 7.30                         & 7.04                         & {\color{blue}9.77 }                        & 7.61
                       \\ & SQ-GAN       & 7.17                         & 7.46                         & 6.89                         & 9.27                         & 7.46
                      \\  & StrokeGAN    & 6.82                         & 7.38                         & 6.99                         & 9.18                         &  7.73
 \\ & StrokeGAN+   & {\color{blue} 7.44}  & {\color{blue} 7.49}  & \textbf{7.76}                &  9.75 & \color{blue}{7.78}                         \\ \hline \hline
     \multirow{7}{*}{\rotatebox{90}{SSIM$\uparrow$}}                    & Pix2Pix      & 0.486                        & {\color{blue} 0.491} & 0.500 & 0.560                        & 0.507
                     \\   & FUNIT        & 0.466                        & 0.467                        & 0.339                        & 0.529                        & \textbf{0.581}
                     \\   & AttentionGAN & \textbf{0.490}               & 0.440                        & {\color{blue} 0.517 }                       & 0.552                        & 0.531
                       \\ & CycleGAN     & 0.462                        & 0.480                        & 0.482                        & {\color{blue} 0.568} & 0.517
                      \\  & SQ-GAN       & 0.471                        & \textbf{0.494}               & 0.479                        & 0.537                        & 0.510
                    \\    & StrokeGAN    & 0.445                        & 0.484                        & 0.495                        & 0.528                        & 0.507
\\ & StrokeGAN+   & {\color{blue} 0.489} &0.490 & \textbf{0.539}               & \textbf{0.571}               & {\color{blue} 0.536}          \\ \hline
\end{tabular}
\end{table}

\subsection{Comparison with State-of-the-art Models}
\label{sc:comparison with sota}

In this section, we implemented fourteen generation tasks including nine printing or handwriting font generation tasks and five calligraphy font generation tasks to demonstrate the effectiveness of the proposed model through comparing with the concerned baselines. The comparison results are shown in Table \ref{Tab:comparison-SOTA-printed} and Table \ref{Tab:comparison-SOTA-calligraphy}.

From Table \ref{Tab:comparison-SOTA-printed} and Table \ref{Tab:comparison-SOTA-calligraphy}, the proposed model achieves the best performance in ten tasks and the second best performance in three tasks among all these fourteen tasks in terms of FID. The outperformance of the proposed model can be also demonstrated by the comparison results in terms of the other three evaluation metrics. Specifically, when compared to Pix2Pix \cite{isola2017image}, a classical model using the paired data, the performance of the proposed model StrokeGAN+ is much better than that of Pix2Pix. This shows that the introduced one-bit stroke encoding and the use of few-shot semi-supervised scheme are helpful to capture the local and global modes of Chinese characters. When compared to these two recent unsupervised models FUNIT \cite{liu2019few} and AttentionGAN \cite{tang2021attentiongan}, the proposed model also outperforms these two models in most of the generation tasks. When concerning the comparison of the performance of these four models, i.e., CycleGAN \cite{chang2018generating}, SQ-GAN \cite{zeng2022Square-BlockGAN}, StrokeGAN \cite{zeng2021strokegan} and StrokeGAN+ suggested in this paper, the performance of StrokeGAN+ is the best over all the generation tasks, and the performance of SQ-GAN and StrokeGAN are comparable and generally better than that of CycleGAN in most of the generation tasks. In particular, it can be observed from Table \ref{Tab:comparison-SOTA-printed} that CycleGAN suffers from the mode collapse issue in three-generation tasks \{fs$\rightarrow$ht, fs$\rightarrow$zk, fs$\rightarrow$ls\}, while the mode collapse can be significantly alleviated by integrating the one-bit stroke encoding as demonstrated by the performance of StrokeGAN, and the performance of StrokeGAN can be further improved by the use of few-shot semi-supervised scheme as shown in the eighth rows of Table \ref{Tab:comparison-SOTA-printed} and Table \ref{Tab:comparison-SOTA-calligraphy}. These verify the effectiveness of the proposed idea, i.e., introducing the one-bit stroke encoding and using the few-shot semi-supervised samples as the supervision information of local and global structure modes of Chinese characters to alleviate the mode collapse issue. Moreover, it can be also observed from Table \ref{Tab:comparison-SOTA-printed} that SQ-GAN still suffers from the mode collapse issue in some generation tasks such as \{fs$\rightarrow$zk\} as shown in the sixth row of Table \ref{Table: mode collapse} and the third row of Figure \ref{Fig:mode collapse and stroke missing fig}, while the mode collapse issue can be addressed by StrokeGAN as well as StrokeGAN+. This also shows the superiority of the proposed idea in reducing mode collapse.

\begin{figure*}[!t]
\begin{center}
	\center{\includegraphics[width=18cm]  {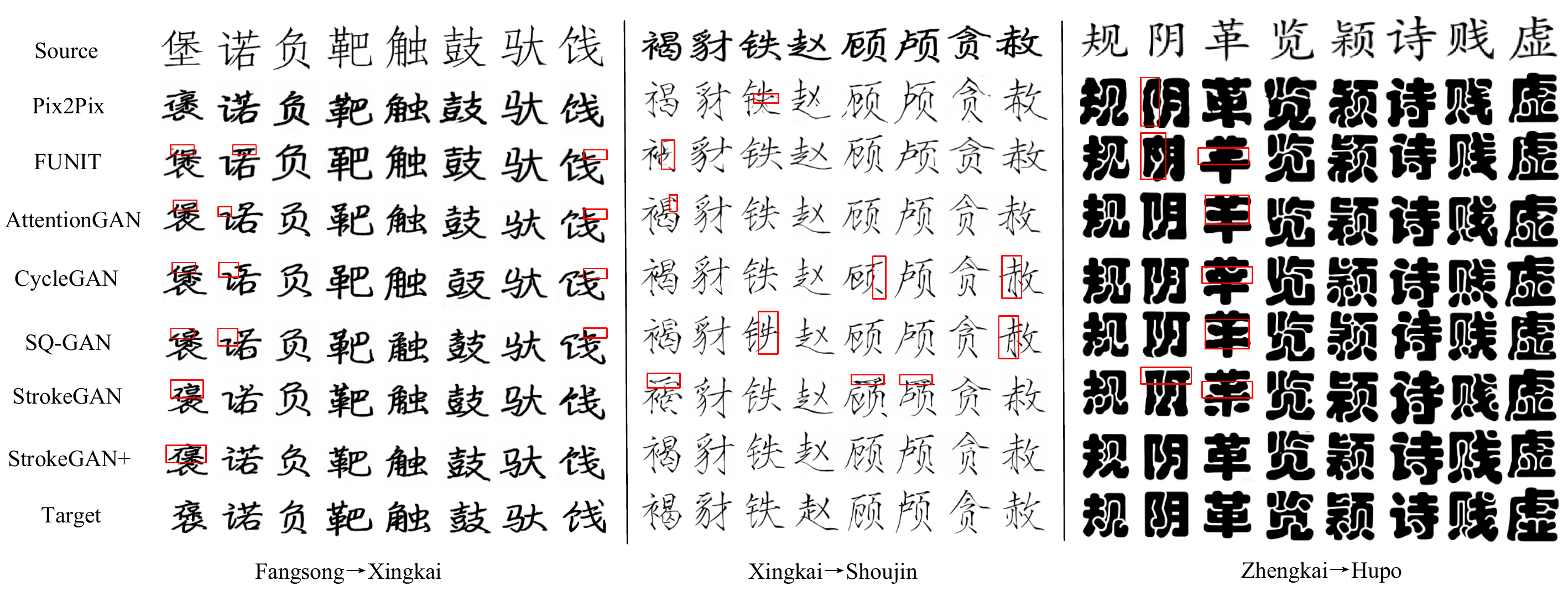}}
	\caption{Some visualization results of generated characters by the proposed model and baselines over three generation tasks.}
	\label{Fig:comparision-sota}
	\end{center}
\end{figure*}

\begin{figure*}[!h]
\begin{center}
	\center{\includegraphics[width=17cm]  {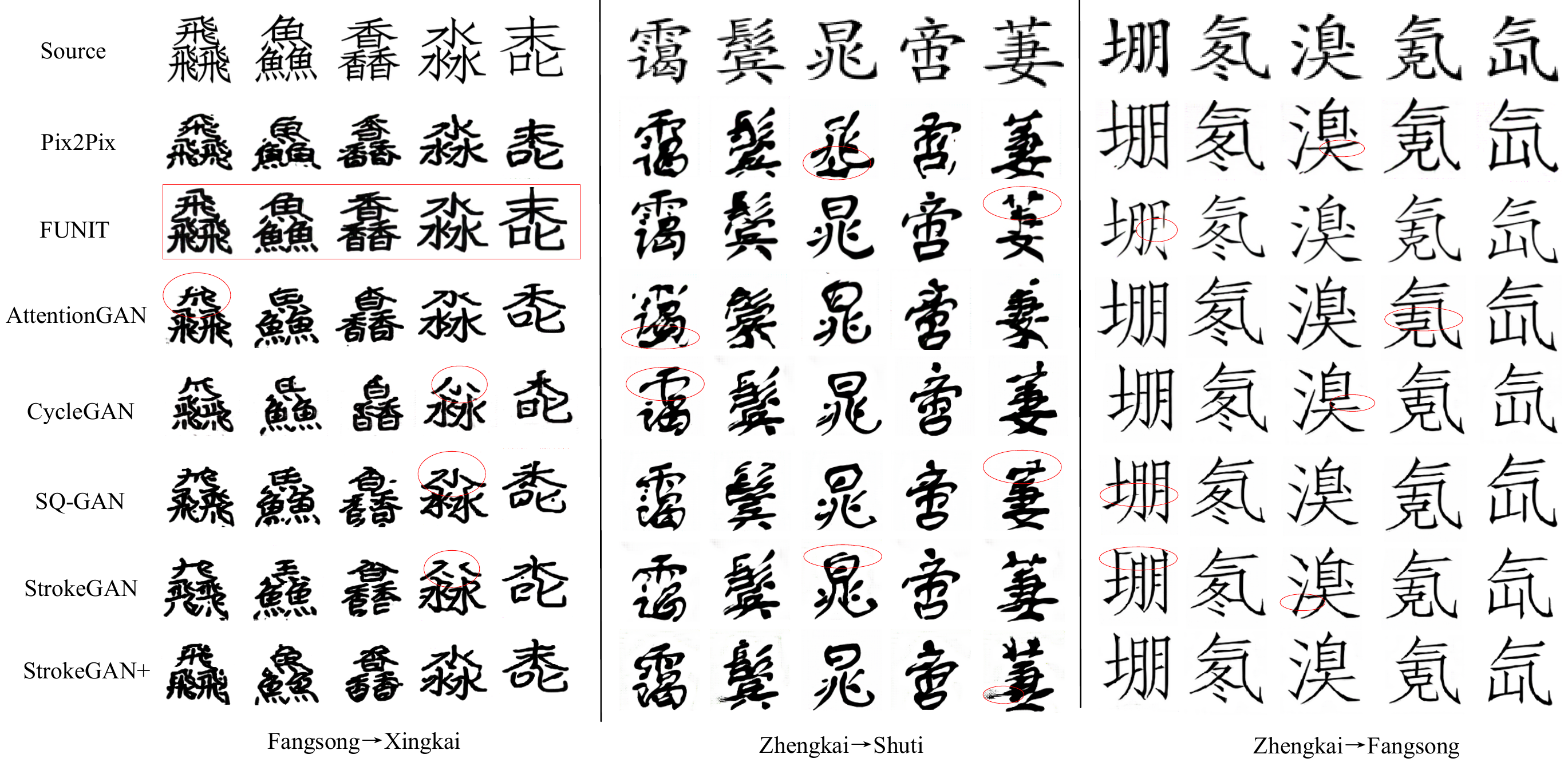}}
	\caption{Comparison on generated characters of the proposed model and baselines for some unseen complex characters.}
	\label{Fig:complex characters}
	\end{center}
\end{figure*}

We also present some visualisation results in Figure \ref{Fig:comparision-sota}, where the generated characters related to three generation tasks are presented. As depicted in Figure \ref{Fig:comparision-sota}, the generated characters of StrokeGAN+ are generally of the highest quality. It can be observed from Figure \ref{Fig:comparision-sota} that there are some flaws in the generated characters of the baselines such as missing strokes and generating some redundant strokes, while this phenomenon can be remarkably alleviated by the proposed model as shown in the eighth row of Figure \ref{Fig:comparision-sota}.


\begin{table*}[!tbh]
\begin{center}
\caption{Comparision on the performance of baselines and their refined versions equipped with the proposed idea in ten printing or handwriting font generation tasks.}
\label{Table:generalization-printed}
\renewcommand\arraystretch{1.2}
\tabcolsep=0.26cm
\begin{tabular}{c|c|c|c|c|c|c|c|c|c|c|c}
\hline
                       & Model         & fs$\rightarrow$xk                  & zk$\rightarrow$hp         & zk$\rightarrow$hw         & fs$\rightarrow$ht         & fs$\rightarrow$zk         & zk$\rightarrow$st         & zk$\rightarrow$dl         & fs$\rightarrow$ls         & hw$\rightarrow$zk         & sj$\rightarrow$xk         \\ \hline \hline
\multirow{4}{*}{FID$\downarrow$}   & AttentionGAN  & \textbf{21.60}           & 58.80          &\textbf{ 37.95}    & 28.31          & 52.68          & 87.17          & 64.67          & \textbf{14.95}  & 57.05          & 25.05          \\
                       & AttentionGAN+ & 35.10           & \textbf{35.36} & 38.20 & \textbf{23.53} & \textbf{26.21} & \textbf{47.88} & \textbf{47.16} & 20.87    & \textbf{32.98} & \textbf{14.07} \\ \cline{2-12}
                       & SQ-GAN        & 49.40                    & \textbf{56.75} & \textbf{59.88} & 32.25          & 269.54         & 71.69          & 65.76          & 26.01                  & 79.75          & 30.29          \\
                       & SQ-GAN+       & \textbf{32.21}  & 68.50 &73.55 & \textbf{27.79} & \textbf{35.42} & \textbf{45.24} & \textbf{43.35} & \textbf{22.12} & \textbf{43.23} & \textbf{25.65} \\ \hline\hline
\multirow{4}{*}{LPIPS$\downarrow$} & AttentionGAN  & 0.209                   & 0.276          & 0.327          & 0.162          & \textbf{0.239} & 0.303          & 0.370    & 0.142    & \textbf{0.305} & 0.318          \\
                       & AttentionGAN+ & \textbf{0.149}  & \textbf{0.189} & \textbf{0.272} & \textbf{0.127} & 0.358          & \textbf{0.188} & \textbf{0.287} & \textbf{0.126}  & 0.381          & \textbf{0.317} \\ \cline{2-12}
                       & SQ-GAN        & 0.219     & 0.297          & \textbf{0.320} & 0.214          & 0.449          & 0.301          & 0.365          & 0.190                & 0.378          & 0.215          \\
                       & SQ-GAN+       & \textbf{0.199}          & \textbf{0.216} & 0.322 & \textbf{0.157} & \textbf{0.132} & \textbf{0.237} & \textbf{0.305} & \textbf{0.146} & \textbf{0.314} & \textbf{0.210}  \\ \hline\hline
\multirow{4}{*}{PSNR$\uparrow$} & AttentionGAN  & 8.17    & 6.76           & 5.71           & 8.90           & 9.18           & 6.91           & 5.17           & \textbf{10.51}         & \textbf{8.08}  & \textbf{6.93}  \\
                       & AttentionGAN+ & \textbf{8.90}   & \textbf{8.12}  & \textbf{6.83}  & \textbf{9.69}  & \textbf{10.41} & \textbf{8.14}  & \textbf{5.89}  & 10.56          & 7.52           & 6.78           \\ \cline{2-12}
                       & SQ-GAN        & \textbf{7.95} &6.68   & \textbf{5.79}  & 7.57           & 7.12           & 7.06           & 5.05           & 8.82                   & 7.47           & 7.46           \\
                       & SQ-GAN+       & 7.81   & \textbf{7.91}  & 5.75  & \textbf{9.08}  & \textbf{10.40} & \textbf{7.83}  & \textbf{5.84}  & \textbf{9.95}   & \textbf{7.85}  & \textbf{7.49}  \\ \hline\hline
\multirow{4}{*}{SSIM$\uparrow$}  & AttentionGAN  & 0.527     & 0.351          & 0.335          & 0.564          & 0.515          & 0.457          & 0.277          & 0.660                   & \textbf{0.448} & \textbf{0.449} \\
                       & AttentionGAN+ & \textbf{0.569}  & \textbf{0.478} & \textbf{0.405} & \textbf{0.619} & \textbf{0.619} & \textbf{0.536} & \textbf{0.331} & \textbf{0.666}  & 0.413          & 0.445          \\ \cline{2-12}
                       & SQ-GAN        & 0.516               & 0.363          & 0.338          & 0.482          & 0.382          & 0.448          & 0.263          & 0.597                & 0.403          & 0.492          \\
                       & SQ-GAN+       & \textbf{0.517}  & \textbf{0.462} & \textbf{0.368} & \textbf{0.589} & \textbf{0.606} & \textbf{0.530} & \textbf{0.333} & \textbf{0.643}  & \textbf{0.447} & \textbf{0.494} \\ \hline
\end{tabular}
\end{center}
\end{table*}

\begin{table}[!tbh]
\renewcommand\arraystretch{1.2}
\caption{Comparision on the performance of baselines and their refined versions equipped with the proposed idea in four calligraphy font generation tasks.}
\label{Table:generalization-calligraphy}
\begin{tabular}{c|c|c|c|c|c}
\hline
                       & Model         & fs$\rightarrow$bs          & fs$\rightarrow$hh          & zk$\rightarrow$cs         & ls$\rightarrow$hc         \\ \hline\hline
\multirow{4}{*}{FID$\downarrow$}   & AttentionGAN  & 56.51          & 92.15          & 40.23         & \textbf{28.66} \\
                       & AttentionGAN+ & \textbf{54.84} & \textbf{52.84} & \textbf{34.08} & 33.16          \\ \cline{2-6}
                       & SQ-GAN        & 55.55          & 59.32          & 58.48          & 45.27          \\
                       & SQ-GAN+       & \textbf{50.87} & \textbf{41.62} & \textbf{31.98} & \textbf{31.43} \\ \hline\hline
\multirow{4}{*}{LPIPS$\downarrow$} & AttentionGAN  & 0.370          & 0.330          & 0.306          & 0.315          \\
                       & AttentionGAN+ & \textbf{0.258} & \textbf{0.311} & \textbf{0.232} & \textbf{0.268} \\ \cline{2-6}
                       & SQ-GAN        & 0.290          & 0.323          & 0.319          & 0.329          \\
                       & SQ-GAN+       & \textbf{0.263} & \textbf{0.311} & \textbf{0.257} & \textbf{0.272} \\ \hline\hline
\multirow{4}{*}{PSNR$\uparrow$}  & AttentionGAN  & 7.41           & 7.41           & 7.35           & 7.72           \\
                       & AttentionGAN+ & \textbf{7.42}  & \textbf{7.52}  & \textbf{7.93}  & \textbf{8.11}  \\ \cline{2-6}
                       & SQ-GAN        & 7.17           & 7.46           & 6.89           & 7.46           \\
                       & SQ-GAN+       & \textbf{7.47}  & \textbf{7.63}  & \textbf{8.05}  & \textbf{8.21}  \\ \hline\hline
\multirow{4}{*}{SSIM$\uparrow$}  & AttentionGAN  & \textbf{0.490} & 0.440          & 0.517          & 0.531          \\
                       & AttentionGAN+ & 0.487          & \textbf{0.496} & \textbf{0.537} & \textbf{0.551} \\ \cline{2-6}
                       & SQ-GAN        & 0.471          & \textbf{0.494} & 0.479          & 0.510          \\
                       & SQ-GAN+       & \textbf{0.495} & 0.482          & \textbf{0.565} & \textbf{0.562} \\ \hline
\end{tabular}
\end{table}
We further test the generalization performance of the proposed model for the generation of some unseen complex Chinese characters, which are excluded from the concerned datasets. Some comparison results are presented in Figure \ref{Fig:complex characters}, where the performance over three generation tasks is presented. From Figure \ref{Fig:complex characters}, the generated characters of the proposed model are generally better than those of baselines. Specifically, there are some flaws such as missing strokes and introducing additional strokes in the generated characters of baselines, while the proposed model rarely suffers from these issues. It can be observed from the third row of Figure \ref{Fig:complex characters} that FUNIT fails in the generation task \{Fangsong $\rightarrow$ Xingkai\} since the styles of generated characters are closer to the style of the source font than that of the target font. These results also demonstrate the good generalization performance of the proposed model.

\begin{figure*}[!t]
\begin{center}
	\center{\includegraphics[width=18cm]  {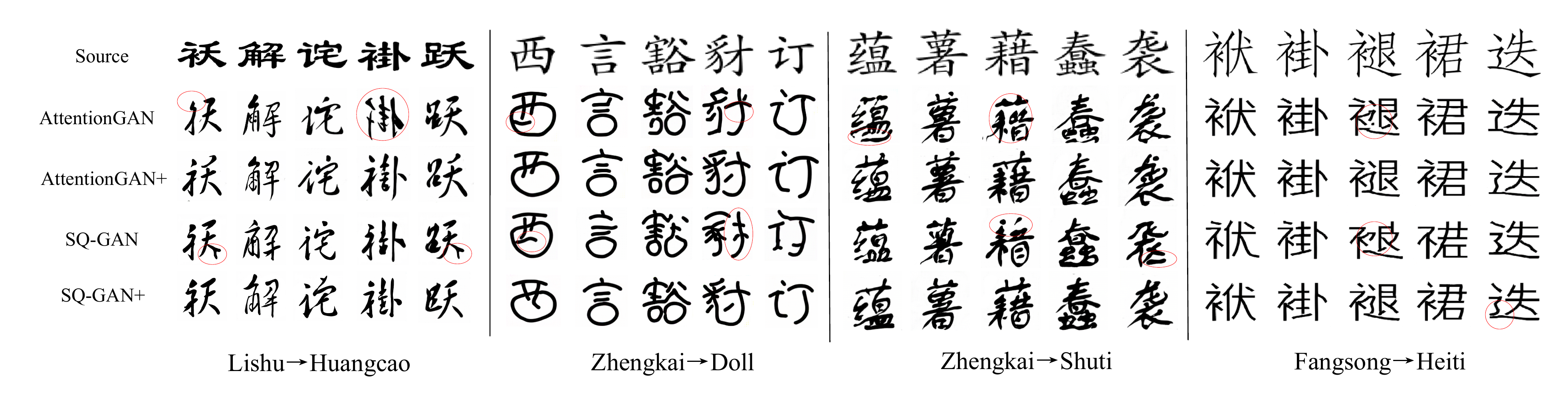}}
	\caption{Comparison on generated characters between two baseslines and their refined versions incorporated with the proposed idea.
 	}
	\label{Fig:comparision-characters-generalization}
	\end{center}
\end{figure*}

\section{Generalizability and Extension}
\label{sec:generalization-extension}
In the following, we implemented a series of experiments to show that the proposed stroke encoding and few-shot semi-supervised scheme can be easily adapted to existing models. The performance of the proposed model in some zero-shot traditional Chinese font generation tasks was also verified at the final of this section.

\subsection{Generalizability of Proposed Idea}
\label{sc:generaliation}
As shown in Figure \ref{Fig:model}, the proposed idea including the stroke encoding and few-shot semi-supervised scheme can be easily adapted to other models besides the CycleGAN model. In this section,  we show the generalizability of proposed ideas over two baselines, i.e., AttentionGAN \cite{tang2021attentiongan} and SQ-GAN \cite{zeng2022Square-BlockGAN}. The refined versions are respectively called \textit{AttentionGAN+} and \textit{SQ-GAN+}. Specifically, for both refined models AttentionGAN+ and SQ-GAN+, only stroke and few-shot semi-supervised modules suggested in this paper are incorporated into the original models similarly in Figure \ref{Fig:model}, where the same 20\% random few-shot semi-supervised scheme is taken as done in StrokeGAN+. We conducted fourteen generation tasks including ten printing or handwriting font generation tasks and four calligraphy font generation tasks to demonstrate the effectiveness of the proposed idea when adapted to AttentionGAN and SQ-GAN. The quantitative comparison results are presented in Table \ref{Table:generalization-printed} and Table \ref{Table:generalization-calligraphy}, and some visualization results are presented in Figure \ref{Fig:comparision-characters-generalization}. From Table \ref{Table:generalization-printed} and Table \ref{Table:generalization-calligraphy}, the refined models equipped with the proposed idea, i.e., AttentionGAN+ and SQ-GAN+ are much better than their original counterparts in most cases. Moreover, as depicted in Figure \ref{Fig:comparision-characters-generalization}, the quality of generated characters is significantly improved over the concerned four generation tasks by combining with the proposed idea.
These show the effectiveness of the proposed idea on enhancing the performance of Chinese font generation models.

\subsection{Effectiveness in Zero-shot Traditional Chinese Font Generation}
\begin{figure*}[!h]
\begin{center}
	\center{\includegraphics[width=18cm]  {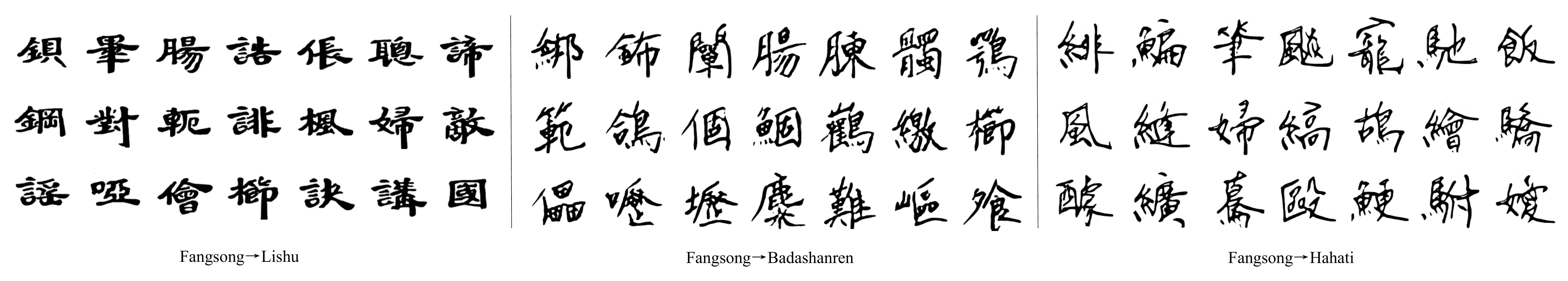}}
	\caption{Generated characters of the proposed model in three zero-shot traditional Chinese font generation tasks. The proposed model was trained over the simplified Chinese characters and tested on the traditional Chinese characters.}
	\label{Fig:zero-shot generation}
	\end{center}
\end{figure*}

In the following, we conducted several experiments to verify the performance of the propose model in the zero-shot traditional Chinese font generation setting. It is known that there are mainly two classes of Chinese characters, i.e., the class of simplified Chinese characters and the class of traditional Chinese characters. Specifically, we trained the proposed model using the simplified Chinese characters and then tested its performance in the generation of the traditional Chinese characters. To verify the performance of the proposed model, we considered three font generation tasks, i.e., \{Fangsong$\rightarrow$Lishu, Fangsong$\rightarrow$Badashanren, Fangsong$\rightarrow$Hahati\}. Some generated characters are presented in Figure \ref{Fig:zero-shot generation}. From Figure \ref{Fig:zero-shot generation}, the proposed model can generate very realistic characters for these traditional Chinese fonts. This show the effectiveness of the proposed model in these zero-shot traditional Chinese font generation settings.



\section{Conclusion}
\label{sc:conclusion}
Existing unsupervised Chinese font generation models may suffer from the mode collapse issue. This paper suggested two kinds of simple but effective supervision information, that is, the one-bit stroke encodings representing the local stroke information and the few-shot semi-supervised samples embodying the global structure information to guide the unsupervised models to capture the local and global structures of Chinese characters. Thus, the mode collapse issue suffered by the known CycleGAN can be effectively alleviated and the performance in Chinese font generation can be significantly improved. The effectiveness of the proposed idea, in particular, the superiority in reducing mode collapse and enhancing the generation performance was demonstrated by numerous experiments. Besides CycleGAN, we showed that our idea can be easily adapted to other deep generative models for the Chinese font generation to further improve their performance. The effectiveness of the proposed model in some zero-shot traditional Chinese font generation settings was also verified.

\ifCLASSOPTIONcompsoc
  \section*{Acknowledgments}
\else
  \section*{Acknowledgment}
\fi
The work of J. Zeng  was supported in part by the National Natural Science Foundation of China [Grant No. 61977038] and by Thousand Talents Plan of Jiangxi Province [Grant No. jxsq2019201124].


\ifCLASSOPTIONcaptionsoff
  \newpage
\fi

\bibliographystyle{IEEEtran}
\bibliography{ref}

\begin{thebibliography}{10}
\providecommand{\url}[1]{#1}
\csname url@samestyle\endcsname
\providecommand{\newblock}{\relax}
\providecommand{\bibinfo}[2]{#2}
\providecommand{\BIBentrySTDinterwordspacing}{\spaceskip=0pt\relax}
\providecommand{\BIBentryALTinterwordstretchfactor}{4}
\providecommand{\BIBentryALTinterwordspacing}{\spaceskip=\fontdimen2\font plus
\BIBentryALTinterwordstretchfactor\fontdimen3\font minus
  \fontdimen4\font\relax}
\providecommand{\BIBforeignlanguage}[2]{{%
\expandafter\ifx\csname l@#1\endcsname\relax
\typeout{** WARNING: IEEEtran.bst: No hyphenation pattern has been}%
\typeout{** loaded for the language `#1'. Using the pattern for}%
\typeout{** the default language instead.}%
\else
\language=\csname l@#1\endcsname
\fi
#2}}
\providecommand{\BIBdecl}{\relax}
\BIBdecl

\bibitem{wang2022aesthetic}
Y.~Wang, G.~Pu, W.~Luo, Y.~Wang, P.~Xiong, H.~Kang, and Z.~Lian, ``Aesthetic
  text logo synthesis via content-aware layout inferring,'' \emph{arXiv
  preprint arXiv:2204.02701}, 2022.

\bibitem{liu2022}
X.~Liu, G.~Meng, J.~Chang, R.~Hu, S.~Xiang, and C.~Pan, ``Decoupled
  representation learning for character glyph synthesis,'' \emph{IEEE
  Transactions on Multimedia}, vol.~24, pp. 1787--1799, 2022.

\bibitem{lin2014font}
\BIBentryALTinterwordspacing
J.-W. Lin, C.-Y. Wang, C.-L. Ting, and R.-I. Chang, ``{Font generation of
  personal handwritten Chinese characters},'' in \emph{Proceedings of the Fifth
  International Conference on Graphic and Image Processing (ICGIP 2013)},
  Y.~Wang, X.~Jiang, M.~Yang, D.~Zhang, and X.~Yi, Eds., vol. 9069,
  International Society for Optics and Photonics.\hskip 1em plus 0.5em minus
  0.4em\relax SPIE, 2014, pp. 352 -- 357. [Online]. Available:
  \url{https://doi.org/10.1117/12.2050128}
\BIBentrySTDinterwordspacing

\bibitem{qian2007towards}
Z.~Qian and D.~Fang, ``Towards chinese calligraphy,'' \emph{Macalester
  International}, vol.~18, no.~1, p.~12, 2007.

\bibitem{chen2011chinese}
T.~Chen, \emph{Chinese Calligraphy}.\hskip 1em plus 0.5em minus 0.4em\relax
  Cambridge University Press, 2011.

\bibitem{goodfellow2014generative}
I.~Goodfellow, J.~Pouget-Abadie, M.~Mirza, B.~Xu, D.~Warde-Farley, S.~Ozair,
  A.~Courville, and Y.~Bengio, ``Generative adversarial nets,'' in
  \emph{Proceedings of the Advances in Neural Information Processing Systems},
  Z.~Ghahramani, M.~Welling, C.~Cortes, N.~Lawrence, and K.~Weinberger, Eds.,
  vol.~27.\hskip 1em plus 0.5em minus 0.4em\relax Curran Associates, Inc.,
  2014.

\bibitem{lyu2017auto}
P.~Lyu, X.~Bai, C.~Yao, Z.~Zhu, T.~Huang, and W.~Liu, ``Auto-encoder guided gan
  for chinese calligraphy synthesis,'' in \emph{Proceedings of the 2017 14th
  IAPR International Conference on Document Analysis and Recognition (ICDAR)},
  vol.~01, 2017, pp. 1095--1100.

\bibitem{tian2017zi2zi}
Y.~Tian, ``zi2zi: Master chinese calligraphy with conditional adversarial
  networks,'' \emph{https://github. com/kaonashi-tyc/zi2zi}, 2017.

\bibitem{wu2020calligan}
S.-J. Wu, C.-Y. Yang, and J.~Y.-j. Hsu, ``Calligan: Style and structure-aware
  chinese calligraphy character generator,'' \emph{arXiv preprint
  arXiv:2005.12500}, 2020.

\bibitem{yuan2022se}
S.~Yuan, R.~Liu, M.~Chen, B.~Chen, Z.~Qiu, and X.~He, ``Se-gan: Skeleton
  enhanced gan-based model for brush handwriting font generation,'' \emph{arXiv
  preprint arXiv:2204.10484}, 2022.

\bibitem{Song2022lffont}
S.~Park, S.~Chun, J.~Cha, B.~Lee, and H.~Shim, ``Few-shot font generation with
  weakly supervised localized representations,'' \emph{IEEE Transactions on
  Pattern Analysis and Machine Intelligence}, pp. 1--17, 2022.

\bibitem{chang2018generating}
B.~Chang, Q.~Zhang, S.~Pan, and L.~Meng, ``Generating handwritten chinese
  characters using cyclegan,'' in \emph{Proceedings of the 2018 IEEE Winter
  Conference on Applications of Computer Vision (WACV)}, 2018, pp. 199--207.

\bibitem{li2019improving}
M.~Li, J.~Wang, Y.~Yang, W.~Huang, and W.~Du, ``Improving gan-based calligraphy
  character generation using graph matching,'' in \emph{2019 IEEE 19th
  International Conference on Software Quality, Reliability and Security
  Companion (QRS-C)}, 2019, pp. 291--295.

\bibitem{jiang2019scfont}
Y.~Jiang, Z.~Lian, Y.~Tang, and J.~Xiao, ``Scfont: Structure-guided chinese
  font generation via deep stacked networks,'' in \emph{Proceedings of the AAAI
  conference on artificial intelligence}, vol.~33, no.~01, 2019, pp.
  4015--4022.

\bibitem{lin2020chinese}
Y.~Lin, H.~Yuan, and L.~Lin, ``Chinese typography transfer model based on
  generative adversarial network,'' in \emph{Proceedings of the 2020 Chinese
  Automation Congress (CAC)}, 2020, pp. 7005--7010.

\bibitem{xie2021dg}
Y.~Xie, X.~Chen, L.~Sun, and Y.~Lu, ``Dg-font: Deformable generative networks
  for unsupervised font generation,'' in \emph{Proceedings of the IEEE/CVF
  Conference on Computer Vision and Pattern Recognition}, 2021, pp. 5130--5140.

\bibitem{isola2017image}
P.~Isola, J.-Y. Zhu, T.~Zhou, and A.~A. Efros, ``Image-to-image translation
  with conditional adversarial networks,'' in \emph{Proceedings of the IEEE
  conference on computer vision and pattern recognition}, 2017, pp. 1125--1134.

\bibitem{zeng2022Square-BlockGAN}
J.~Zeng, Q.~Chen, and M.~Wang, ``Self-supervised chinese font generation based
  on square-block transformation (in chinese),'' \emph{Sci Sin Inform},
  vol.~52, no.~1, pp. 145--159, 2022.

\bibitem{zeng2021strokegan}
J.~Proceedings of~the Zeng, Q.~Chen, Y.~Liu, M.~Wang, and Y.~Yao, ``Strokegan:
  Reducing mode collapse in chinese font generation via stroke encoding,'' in
  \emph{Proceedings of AAAI}, vol.~3, 2021.

\bibitem{zhu2017unpaired}
J.-Y. Zhu, T.~Park, P.~Isola, and A.~A. Efros, ``Unpaired image-to-image
  translation using cycle-consistent adversarial networks,'' in
  \emph{Proceedings of the Proceedings of the IEEE international conference on
  computer vision}, 2017, pp. 2223--2232.

\bibitem{lian2012automatic}
Z.~Lian and J.~Xiao, ``Automatic shape morphing for chinese characters,'' in
  \emph{Proceedings of the SIGGRAPH Asia 2012 Technical Briefs}, 2012, pp.
  1--4.

\bibitem{liu2012automatic}
P.~Liu, S.~Xu, and S.~Lin, ``Automatic generation of personalized chinese
  handwriting characters,'' in \emph{Proceedings of the 2012 Fourth
  International Conference on Digital Home}.\hskip 1em plus 0.5em minus
  0.4em\relax IEEE, 2012, pp. 109--116.

\bibitem{lin2015complete}
J.-W. Lin, C.-Y. Hong, R.-I. Chang, Y.-C. Wang, S.-Y. Lin, and J.-M. Ho,
  ``Complete font generation of chinese characters in personal handwriting
  style,'' in \emph{2015 IEEE 34th International Performance Computing and
  Communications Conference (IPCCC)}, 2015, pp. 1--5.

\bibitem{lin2019font}
X.~Lin, J.~Li, H.~Zeng, and R.~Ji, ``Font generation based on least squares
  conditional generative adversarial nets,'' \emph{Multimedia Tools and
  Applications}, vol.~78, no.~1, pp. 783--797, 2019.

\bibitem{wen2021handwritten}
C.~Wen, Y.~Pan, J.~Chang, Y.~Zhang, S.~Chen, Y.~Wang, M.~Han, and Q.~Tian,
  ``Handwritten chinese font generation with collaborative stroke refinement,''
  in \emph{Proceedings of the IEEE/CVF Winter Conference on Applications of
  Computer Vision}, 2021, pp. 3882--3891.

\bibitem{gao2020gan}
Y.~Gao and J.~Wu, ``Gan-based unpaired chinese character image translation via
  skeleton transformation and stroke rendering,'' in \emph{proceedings of the
  AAAI conference on artificial intelligence}, vol.~34, no.~01, 2020, pp.
  646--653.

\bibitem{qin2022disentangled}
M.~Qin, Z.~Zhang, and X.~Zhou, ``Disentangled representation learning gans for
  generalized and stable font fusion network,'' \emph{IET Image Processing},
  vol.~16, no.~2, pp. 393--406, 2022.

\bibitem{liu2019few}
M.-Y. Liu, X.~Huang, A.~Mallya, T.~Karras, T.~Aila, J.~Lehtinen, and J.~Kautz,
  ``Few-shot unsupervised image-to-image translation,'' in \emph{Proceedings of
  the IEEE/CVF International Conference on Computer Vision}, 2019, pp.
  10\,551--10\,560.

\bibitem{tang2021attentiongan}
H.~Tang, H.~Liu, D.~Xu, P.~H.~S. Torr, and N.~Sebe, ``Attentiongan: Unpaired
  image-to-image translation using attention-guided generative adversarial
  networks,'' \emph{IEEE Transactions on Neural Networks and Learning Systems},
  pp. 1--16, 2021.

\bibitem{ioffe2015batch}
S.~Ioffe and C.~Szegedy, ``Batch normalization: Accelerating deep network
  training by reducing internal covariate shift,'' in \emph{Proceedings of the
  International conference on machine learning}.\hskip 1em plus 0.5em minus
  0.4em\relax PMLR, 2015, pp. 448--456.

\bibitem{kingma2014adam}
D.~P. Kingma and J.~Ba, ``Adam: A method for stochastic optimization,''
  \emph{arXiv preprint arXiv:1412.6980}, 2014.

\bibitem{heusel2017gans}
M.~Heusel, H.~Ramsauer, T.~Unterthiner, B.~Nessler, and S.~Hochreiter, ``Gans
  trained by a two time-scale update rule converge to a local nash
  equilibrium,'' \emph{Proceedings of the Advances in neural information
  processing systems}, vol.~30, 2017.

\bibitem{zhang2018unreasonable}
R.~Zhang, P.~Isola, A.~A. Efros, E.~Shechtman, and O.~Wang, ``The unreasonable
  effectiveness of deep features as a perceptual metric,'' in \emph{Proceedings
  of the IEEE conference on computer vision and pattern recognition}, 2018, pp.
  586--595.

\bibitem{Wang2004-SSIM}
Z.~Wang, A.~C. Bovik, H.~R. Sheikh, and E.~P. Simoncelli, ``Image quality
  assessment: From error visibility to structural similarity,'' \emph{IEEE
  Transactions on Image Processing}, vol.~13, no.~4, pp. 600--612, 2004.

\end{thebibliography}
\end{document}